%% file: main.tex
\documentclass[acmsmall,screen]{acmart}

\usepackage{amsmath}

\usepackage{amssymb}
\usepackage{hyperref}
\usepackage{graphicx}
\usepackage{textcomp}
\usepackage{array}
\usepackage{xcolor}
\usepackage{tcolorbox}
\usepackage{colortbl}
\usepackage{svg}
\usepackage[linesnumbered,ruled,vlined]{algorithm2e}
\usepackage{algpseudocode}
\usepackage{multirow}
\usepackage{booktabs}
\usepackage{threeparttable}
\usepackage{enumitem}
\usepackage{caption}
\usepackage{listings}
\usepackage{multicol}
\usepackage{wrapfig}
\usepackage{pdfpages}
\usepackage{tikz}
\newcommand{\mybox}[1]{%
  \begin{tcolorbox}[colback=white,colframe=black,lowerbox=invisible,savelowerto=\jobname_ex.tex]
    #1
  \end{tcolorbox}
}

\AtBeginDocument{%
  }

\setcopyright{acmlicensed}
\acmDOI{10.1145/3715735}
\acmYear{2025}
\acmJournal{PACMSE}
\acmVolume{2}
\acmNumber{FSE}
\acmArticle{FSE020}
\acmMonth{7}
\received{2024-09-13}
\received[accepted]{2025-01-14}

\acmConference[Conference acronym 'XX]{Make sure to enter the correct
  conference title from your rights confirmation email}{June 03--05,
  2018}{Woodstock, NY}

\begin{document}


\title{Hallucination Detection in Large Language Models with Metamorphic Relations}

\author{Borui Yang}
\orcid{0009-0009-3482-7667}
\affiliation{%
  \institution{King's College London}
  \city{London}
  \country{United Kingdom}
}
\email{zbybr@bupt.edu.cn}

\author{Md Afif Al Mamun}
\orcid{0000-0002-9319-3483}
\affiliation{%
  \institution{University of Calgary}
  \city{Calgary}
  \country{Canada}
}
\email{afif.mamun@ucalgary.ca}

\author{Jie M. Zhang}
\orcid{0000-0003-0481-7264}
\affiliation{%
  \institution{King's College London}
  \city{London}
  \country{United Kingdom}
}
\email{jie.zhang@kcl.ac.uk}

\author{Gias Uddin}
\orcid{0000-0003-1376-095X}
\affiliation{%
  \institution{York University}
  \city{Toronto}
  \country{Canada}
}
\email{guddin@yorku.ca}

\begin{abstract}
    Large Language Models (LLMs) are prone to hallucinations, e.g.,   factually incorrect information, in their responses. These hallucinations present challenges for LLM-based applications that demand high factual accuracy.  Existing hallucination detection methods primarily depend on external resources, which can suffer from issues such as low availability, incomplete coverage, privacy concerns, high latency, low reliability, and poor scalability. There are also methods depending on output probabilities, which are often inaccessible for closed-source LLMs like GPT models. This paper presents MetaQA, a self-contained hallucination detection approach that leverages metamorphic relation and prompt mutation. 
    Unlike existing methods, MetaQA operates without any external resources and is compatible with both open-source and closed-source LLMs. 
    MetaQA is based on the hypothesis that if an LLM's response is a hallucination, the designed metamorphic relations will be violated. 
We compare MetaQA with the state-of-the-art zero-resource hallucination detection method, SelfCheckGPT, across multiple datasets, and on two open-source and two closed-source LLMs.
Our results reveal that    MetaQA outperforms SelfCheckGPT in terms of precision, recall, and f1 score.
     For the four LLMs we study, MetaQA outperforms SelfCheckGPT with a superiority margin ranging from 0.041 - 0.113 (for precision), 0.143 - 0.430 (for recall), and 0.154 - 0.368 (for F1-score).
     For instance, with Mistral-7B, MetaQA achieves an average
F1-score of 0.435, compared to SelfCheckGPT’s F1-score of 0.205, representing an improvement rate of
112.2\%.
     MetaQA also demonstrates superiority across all different categories of questions. 

\end{abstract}



\maketitle

\section{Introduction}

\begin{wrapfigure}[9]{r}{0.45\linewidth}
    \vspace{-1em}
    \centering
    \includegraphics[width=0.9\linewidth]{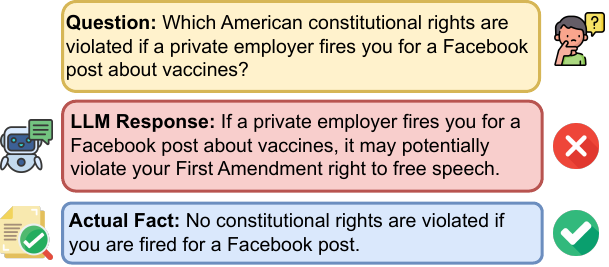}
    \vspace{-3mm}
    \caption{An example of a hallucinated output generated by ChatGPT in the legal domain.}
    \label{example}
\end{wrapfigure}

Large Language Models (LLMs) like GPT-4 \cite{openai2023gpt} have revolutionized language processing by generating fluent and organized responses for various applications, such as drafting reports and summarization systems \cite{moramarco2021preliminary, lai2022exploration, zhang2023concepteva}. 
However, LLMs are prone to generating hallucinations—coherent but factually incorrect or irrelevant outputs,
such as non-factual responses in question-answering context \cite{huang2023survey, xu2024hallucination}.
This tendency poses significant challenges to the reliability of LLMs, undermining the effectiveness of LLMs in applications requiring high factual accuracy. 
Fact-conflicting hallucinations, where LLMs produce content that contradicts established facts, are particularly concerning as they can mislead users lacking expertise on the topic, leading to significant confusion and eroding the trust essential for various LLM applications. Figure \ref{example} illustrates a scenario where a legal question is posed to ChatGPT, which generates a response containing hallucinations. Without proper detection, such inaccuracies could have serious consequences for individuals who lack expertise in the legal field.

Many approaches have been introduced to detect or mitigate hallucination of LLMs that compare responses to factual information, often relying on databases or search engines \cite{guo2022survey, augenstein2019multifc, Hanselowski2019ARA, Atanasova2020GeneratingFC, 202307.1723, huo2023retrieving}. 
However, relying on external resources for hallucination detection often 
 limits the scope to specific domains where a comprehensive database does not exist.
Moreover, hallucinations are observed across a wide range of tasks that extend beyond simple fact verification~\cite{kryscinski2019evaluating}.
Other works use token-level information such as token confidence and entropy \cite{varshney2023stitch, yao2023llm}, which is often inaccessible in closed-source models. 
To tackle these issues, Manakul et al. proposed SelfcheckGPT \cite{manakul2023selfcheckgpt}, a self-contained approach to addressing hallucination.
However, for the sample generation process in SelfCheckGPT, the LLM tends to produce samples that are identical to the original response, which significantly impacts its performance in hallucination detection.


This paper introduces MetaQA, a novel zero-resource technique that employs Metamorphic Relations (MRs) to detect hallucinations in LLM responses. MetaQA uses MRs to generate response mutations and verifies these mutations against expected outcomes to identify inconsistencies. Acting as a test oracle—similar to mechanisms in software testing, MetaQA ensures reliable factual accuracy checks of the LLM outputs without requiring additional agents. This method is advantageous as it relies solely on the LLM itself. MetaQA is applicable to both open-source and closed-source LLMs. MetaQA can be used without the need for intermediate processes or external tools.

We evaluate MetaQA using three datasets—TruthfulQA, HotpotQA, and FreshQA — the widely studied benchmarks in hallucination evaluation, across four LLMs: GPT-4, GPT-3.5, Llama3, and Mistral. 
Our results reveal that 
MetaQA consistently outperforms SelfCheckGPT in terms of precision,
recall, and f1 score on all the four LLMs we study. In particular, for the four LLMs, MetaQA outperforms SelfCheckGPT with a superiority margin ranging from 0.154 to 0.368 in terms of F1-score.
Our ablation studies also show that
MetaQA has considerable stability across multiple runs, and has better performance with 
lower temperatures.

Our major contributions include:

\begin{enumerate}[leftmargin=20pt]
    \item To the best of our knowledge, we are the first to apply synonym and antonym-based metamorphic relations to detect hallucination LLMs responses.
    \item We improve the TruthfulQA \cite{lin2021truthfulqa} benchmark by updating 238 questions with new correct answers. We share the improved benchmark, which we name \textit{TruthfulQA-Enhanced}. This new benchmark can support more accurate hallucination detection research.
    \item We conduct a large-scale evaluation of MetaQA on TruthfulQA-Enhanced, FreshQA, and HotpotQA datasets, showing superior performance over the baseline method SelfCheckGPT.
\end{enumerate}

\section{Preliminaries}
\subsection{Motivating Example}\label{sec:motivation}

\begin{figure*}[t]
    \centering
    \includegraphics[width=.9\textwidth]{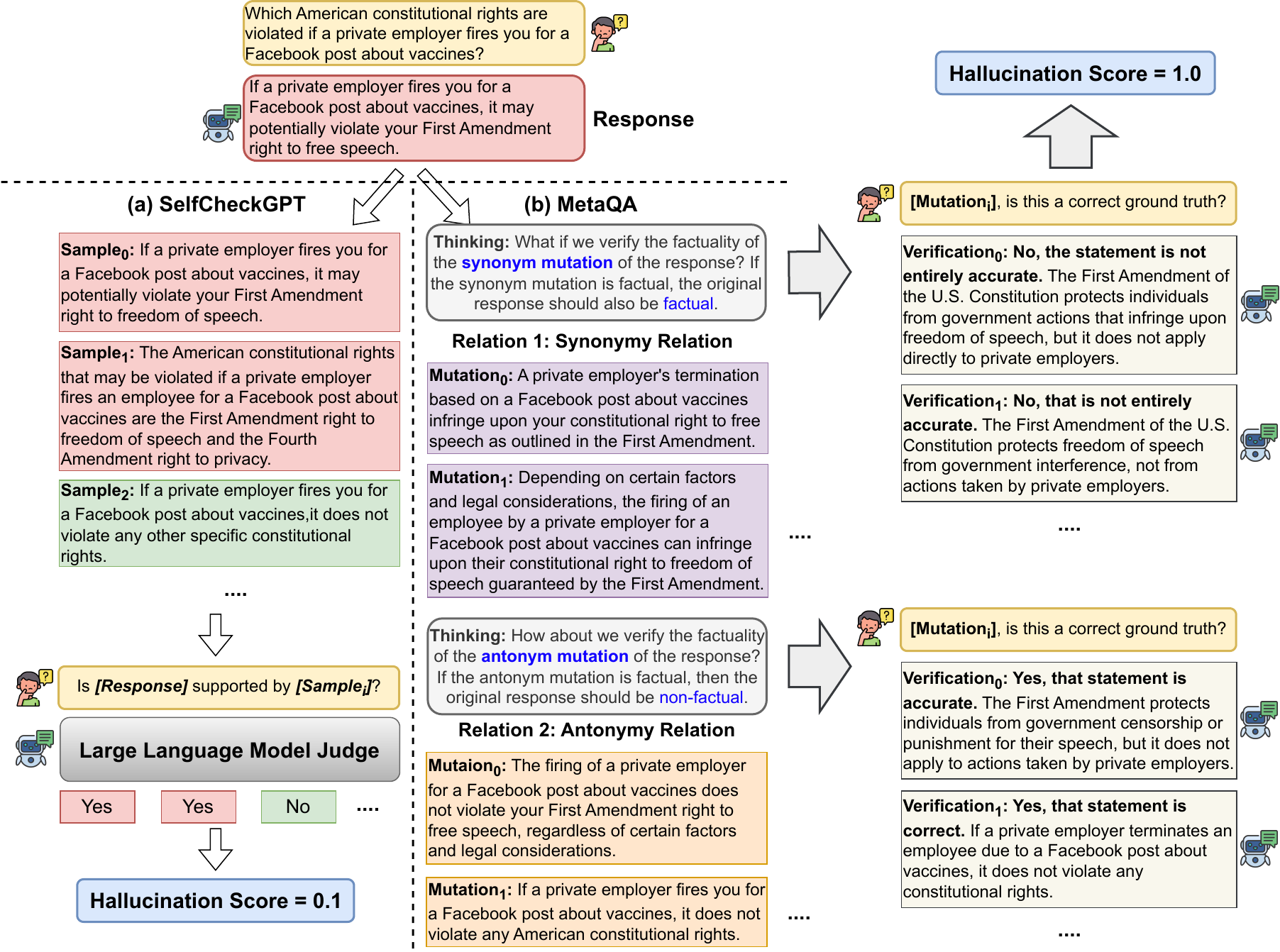}
    \caption{A sample of our motivating idea on MetaQA and a simple comparison where SelfCheckGPT fails to detect hallucination.}
    \label{fig:motivate}
\end{figure*}

To further motivate our approach, we reexamine the hallucinated example in Figure \ref{example} using SelfCheckGPT, the method closely aligned with ours, but which fails to detect the hallucination. SelfCheckGPT operates by generating \( N \) response samples for the same query and comparing their semantic similarity to the base response. It calculates a \textit{hallucination score} based on whether these samples support the factual content of the base response. Figure \ref{fig:motivate} shows that when the LLM was repeatedly prompted with the same query using SelfCheckGPT, it consistently generated similar hallucinated responses most of the time. This led to a hallucination score of 0.1, as the generated samples mostly remained consistent with the base response, reinforcing the incorrect information. 

In contrast, MetaQA applies Metamorphic Relations (MR)~\cite{chen2018metamorphic, zhang2014search} to introduce controlled mutations by generating a set of mutations from the base response. Formally, let \( B \) be the base response. MetaQA defines a function \( f(B) \) using the same LLM that produces two sets of mutations: a set of synonymous mutations \( S = f_s(B) = \{ s_0, s_1, s_2, \dots, s_N \} \) and a set of antonymous mutations \( A = f_a(B) = \{ a_0, a_1, a_2, \dots, a_M \} \). For each mutation, the system verifies whether the transformation remains factually consistent. This verification process assigns a hallucination score based on the factual alignment of both mutation sets \( S \) and \( A \). The advantage of MR over repeated prompts is that MR introduces changes to the input that make LLMs reveal inconsistencies more effectively, particularly when hallucinations are present \cite{hyun2024metal}. While repeated prompts often produce consistent outputs—even if the base response contains hallucinations—MR's varied mutations expose these inconsistencies by prompting the LLM to produce divergent responses. Additionally, independently verifying each mutation reduces bias from previous outputs, resulting in a more accurate and reliable detection of hallucinations \cite{dhuliawala2023chain}. Revisiting the example shown in Figure \ref{fig:motivate}, we observe that while SelfCheckGPT computed a very low hallucination score, our approach—which uses various MRs and validates each mutation individually—resulted in a high hallucination score of $1.0$ as none of the mutations were validated as a fact. This indicates a greater likelihood of detecting hallucinations in this case.

\begin{wrapfigure}[14]{r}{0.5\linewidth}
    \centering
    \includegraphics[width=0.9\linewidth]{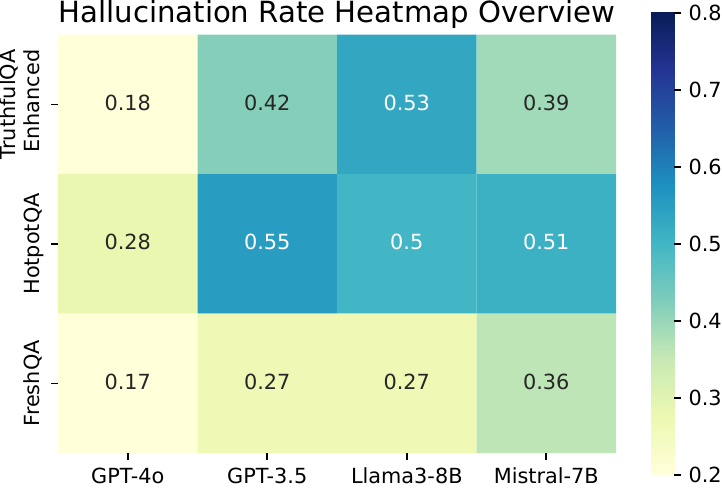}
    \caption{\textbf{}Overall Hallucination Rate of Different LLMs. }
    \label{overallhalu}
\end{wrapfigure}

\subsection{LLM Hallucination Severity}

Despite recent advancements in Large Language Models (LLMs), they still suffer significantly from the issue of hallucinations.
Figure~\ref{overallhalu} 
presents a heatmap depicting the hallucination rates of various experimental Large Language Models (LLMs) across three different datasets: TruthfulQA Enhanced, HotpotQA, and FreshQA. The hallucination rates are represented by the numerical values within each cell, where lower values indicate a lower rate of hallucinations and higher values indicate a higher rate.
We can observe that current LLMs have a significant tendency to produce hallucinations in question answering, with rates ranging from 17\% to 55\%. GPT-3.5, Llama3, and Mistral all exhibited high hallucination rates, particularly in the experiments on the HotpotQA dataset, with rates of 55\%, 50\%, and 51\% respectively. Only GPT-4 demonstrated a significantly lower hallucination rate, with rates of 18\% and 17\% on the TruthfulQA-Enhanced and FreshQA datasets, and 28\% on the HotpotQA dataset.

\subsection{LLM Hallucination Types} \textbf{Hallucination of LLMs} has been extensively studied in the realm of Natural Language Processing (NLP). Hallucination typically refers to a phenomenon where the generated content appears nonsensical or unfaithful to the provided source content \cite{ji2023survey}. Generally, hallucination in natural language generation tasks can be categorized into three primary types \cite{yao2024survey, zhang2023siren}, as detailed below:
\begin{itemize}[leftmargin=10pt]
    \item \textbf{Input-Conflicting Hallucination}: This type occurs when LLM generates outputs that are inconsistent with the user's input. Typically, such inconsistencies can present themselves in two primary ways: the LLM's response might conflict with the user's task instructions, suggesting a misinterpretation of the intent, or the generated output might contradict the task input, similar to common issues in machine translation, or summarization \cite{li2022faithfulness}. 
    \item \textbf{Context-Conflicting Hallucination}: This arises when the output of the LLM contradicts the contextual information provided by the user. Such inconsistencies often occur in lengthy or multi-turn interactions, where the model may lose track of the context or fail to maintain consistency throughout the conversation.
    \item \textbf{Fact-Conflicting Hallucination} occur when generated content contradicts established world knowledge \cite{chen2023complex, chern2023factool, min2023factscore}. Several factors throughout the life-cycle of an LLM, including the training dataset, pre-training procedures, and the inference process, can contribute to such type of hallucination. For instance, as shown in Figure \ref{example}, an LLM might present inaccurate information in response to a user query, potentially misleading users who lack expertise on the topic.
\end{itemize}
The dissemination of incorrect information through misleading answers can have serious consequences. Therefore, in this paper, we focus on the detection of fact-conflicting hallucinations. Such hallucinations can significantly undermine trust in the generated content, particularly when users lack knowledge of the correct ground truth.

\subsection{Metamorphic Relation (MR)} A metamorphic relationship $R$ is a necessary property between the set of inputs $X = <x_1, x_2, ..., x_n>$ and their corresponding outputs $Y = <f(x_1), f(x_2), ... f(x_n)>$ from a given function $f$ and is denoted as $R\subseteq X^n \times Y^n$ or simply as $R(x_1, x_2, ..., x_n, f(x_1), f(x_2), ... f(x_n))$~\cite{chen2018metamorphic,zhang2014search}. Metamorphic relations can be applied independently or in conjunction with other static and dynamic software analysis techniques, such as formal proofs and debugging. Essentially, an MR is a property or constraint that the output of a system should satisfy when specific transformations or modifications are applied to its input. If the system's output changes inappropriately in response to these transformations, it may indicate that the system is producing incorrect or unreliable results.

In the context of Natural Language Generation, combining Metamorphic Relations provides a robust method for detecting hallucination issues in LLM. Applying MRs allows verification of whether the model maintains expected semantic relationships when faced with various input transformations.
 nor it indicate the limitation of the existing approach (i.e. SelfcheckGPT)

\section{MetaQA Methodology}

\begin{figure}[!htbp]
    \centering
    \includegraphics[width=1\linewidth]{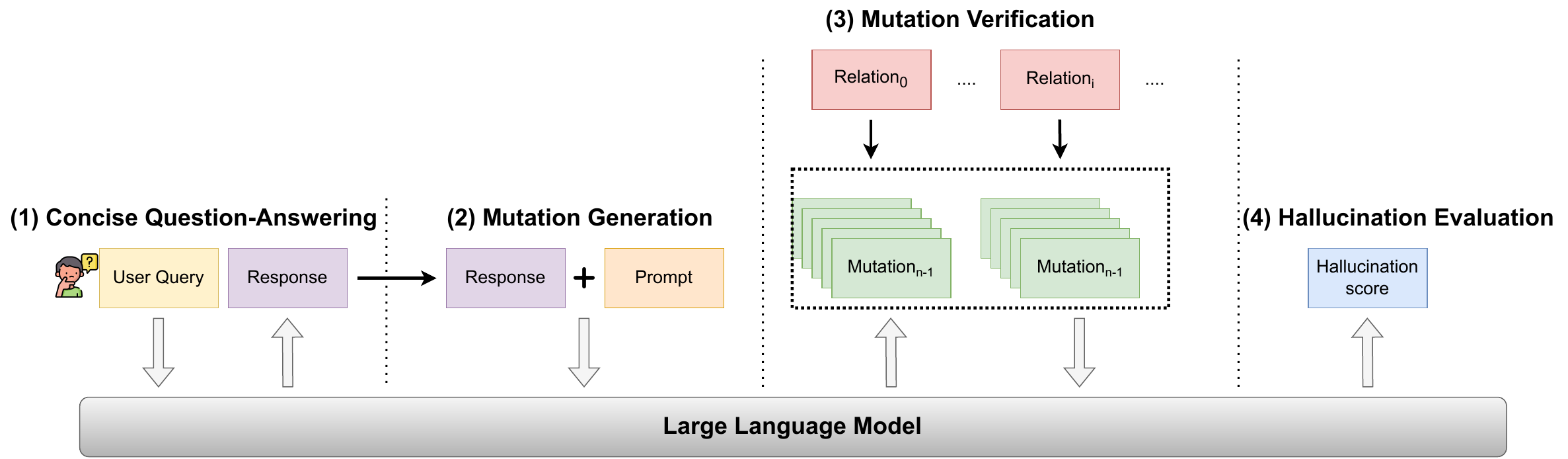}
    \caption{The workflow of MetaQA}
    \label{fig:workflow}
\end{figure}

MetaQA is a hallucination detection framework that focuses on detecting factually conflicting outputs. MetaQA does this by comparing multiple responses for a given question, where follow-up questions to the given question are generated by using MRs. The MetaQA framework is structured into five steps (see Figure \ref{fig:workflow}): \textit{(1) Concise Question-Answering, (2) Mutation Generation, (3) Mutation Verification, and (4) Hallucination Evaluation}. Algorithm \ref{alg:mutation-generation} outlines the complete process from generating mutations based on the initial LLM response to calculating the hallucination score.

\input{algo/algo}

\subsection{Step 1. Concise Question-Answering}
We guide LLM to produce concise and fact-based answers in response to an input question. In typical interactions, LLMs often generate detailed and lengthy explanations. However, within the MetaQA framework, excessive verbosity can hinder downstream modules involved in tasks such as mutation generation and verification. To address this, we employ system prompts that instruct the LLM to generate brief, contextually grounded answers. This ensures that responses remain precise. Recent studies have shown that LLMs are effective in evaluating the consistency of information between lengthy documents and concise summaries, even in zero-shot settings \cite{luo2023chatgpt}. Therefore, in addition to the standard question-answering process, we instruct the model to summarize its responses into short, accurate sentences. 

\subsection{Step 2. Mutation Generation}
For a given response (we call it \textit{base} response)  to a question from Step 1, we create multiple distinct high-quality mutations to the response. Each mutation is produced as a follow-up question to the base response. MetaQA employs a prompt-based approach that exclusively uses the LLM to generate mutations\footnote{
We do not adopt traditional synonym- and antonym-based mutation generation methods~\cite{sun2020automatic,9000651} because LLMs have consistently demonstrated superior performance over traditional NLP techniques across almost all NLP tasks~\cite{qin2024large,foster2025mutation}.}. Thus, for potentially overly brief responses from the LLM, this method will generate complete and semantically accurate mutations based on the context of the question. A mutation is denoted as being generated in the following manner: 
\begin{equation}
    mutation = f(subject, relation)
\end{equation}
where $subject$ is constructed by the question-response pair. In Algorithm \ref{alg:mutation-generation}, the steps for generating mutations based on a predefined relation are detailed in lines \ref{line:init_mt_generation} to \ref{line:end_mt_generation}. These steps involve using a predefined prompt template for each type of metamorphic relationship. We will explore these relationships in greater detail in the following sections.


We deploy two general types of reasoning rules prevalently adopted in several literature \cite{abboud2020boxe, liang2024survey, ren2020beta, tian2022knowledge, zhou2019completing}, the types of metamorphic relations are detailed as follows:

\subsubsection{Relation 1: Synonymy Relation}

A synonymous mutation refers to a variation of the response generated by the LLM that maintains the same semantic meaning as the original response. For example, a general structure of the response sentence typically follows a specific order: ($subject$, $verb$, $object$). In the synonymy relation, the following types are included:
\begin{itemize}[leftmargin=10pt]
    \item \textbf{Lexical Substitution}: The subject and object in a sentence remain unchanged, semantic consistency is maintained by replacing other parts of the sentence with synonyms.
    \begin{equation}
        R(sub, verb_{1}, obj) \Rightarrow R'(sub, verb_{2}, obj)
    \end{equation}
    where $verb_{2}$ should share the close meaning to $verb_{1}$, plus fits the new structure.
    \item \textbf{Inversion}: In an inverse relationship, the subject and object can be reversely linked through a variant of the original relation, as demonstrated below:
    \begin{equation}
        R(sub, verb, obj) \Rightarrow R'(obj, verb', sub)
    \end{equation}
    where $verb'$ is the modified $verb$ that fits the new structure.
\end{itemize}
In this context, evaluating the synonymous mutation of a response aims to determine whether such mutation, which maintains semantic equivalence, is classified as factual or non-factual, thereby assessing whether the initial response manifests hallucination.

\subsubsection{Relation 2: Antonymy Relation}

In the domain of antonymy, given a question-response pair, we can verify the authenticity of the antonymous mutation, which conveys an opposite semantic meaning to the original response. With this relation, for a response $R$ without hallucination, its antonymous mutation $\overline{R}$ should not be considered as the correct ground truth. An antonymous mutation can be generated by the formula below:
\begin{equation}
    R(sub, verb, obj) \Rightarrow \overline{R}(sub, \neg{verb}, obj)
\end{equation}

As illustrated in Table \ref{tab1}, before initiating our interaction with LLM, we predefined specific instructions and prompt templates, requesting the model to use its inherent knowledge and inferential capabilities to deliver accurate and specific content as per our requirements. The primary aim is to ensure LLM provides easily analyzable responses by using standardized prompts and instructions.

\begin{table}[!htbp]
    \centering
    \caption{Templates for generating synonym and antonym mutations in the mutation process.}
    \resizebox{\textwidth}{!}{
        \begin{tabular}{p{4.19em}llllll|p{4.19em}llllll}
        \toprule
        \multicolumn{7}{c|}{\textbf{Synonym Mutation}} & \multicolumn{7}{c}{\textbf{Antonym Mutation}} \\
        \midrule
        \multicolumn{7}{l|}{\textbf{Instruction}: Generate synonym mutations of the base response} & \multicolumn{7}{l}{\textbf{Instruction}: Generate antonym mutations of the base response} \\
        \midrule
        \multicolumn{7}{p{29.33em}|}{\textbf{Query}: Generate synonym mutations of the answer based on the context of the question and return a numbered list to me.\newline{}Do not add any information that's not provided in the answer nor asked by the question. Make sure the generated synonyms are meaningful sentences.\newline{}For example:\newline{}Question: What is the most popular sport in Japan?\newline{}Answer: Baseball is the most popular sport in Japan.\newline{}Mutations:\newline{}1. Japan holds baseball as its most widely embraced sport.\newline{}2. The sport with the highest popularity in Japan is baseball.\newline{}3. Baseball reigns as Japan's most favored sport among the populace.\newline{}Notice how the full context is included in each generated synonym. If\newline{}you generated just `baseball', it would not make a meaningful sentence.} & \multicolumn{7}{p{29.33em}}{\textbf{Query}: Generate negations of the answer based on the context of the question and return a numbered list to me.\newline{}Do not add any information that's not provided in the answer nor asked by the question. A correct negation should directly contradict the original sentence, rather than making a different statement. Make sure the generated antonyms are meaningful sentences.\newline{}For example:\newline{}Question: What is the most popular sport in Japan?\newline{}Answer: Baseball is the most popular sport in Japan.\newline{}Mutations:\newline{}1. The most popular sport in Japan is not baseball.\newline{}2. Baseball is not the most popular sport in Japan.\newline{}3. Japan does not consider baseball as the most popular sport.\newline{}Be careful about double negations which make the sentence semantically same to the provided one. The context of the question is really important. Notice how the negations are meaningful sentences in the example. You should negate the meaning of the sentence based on the question.} \\
        \bottomrule
        \end{tabular}%
        }
    \label{tab1}%
  \end{table}%

\subsection{Step 3. Mutation Verification}
Based on the generated mutations, MetaQA prepares test cases and obtains verification results through a straightforward yet effective question-answering process with LLMs. The Mutation Verification module within MetaQA serves as a specialized component for validating mutations derived from the original question-response pairs. This module independently verifies each mutation with respect to the MR used to generate the mutation (Line \ref{line:mt_verification} in Algorithm \ref{alg:mutation-generation}). It employs a prompt-based method that allows the LLM itself to rigorously assess whether the mutations conform to established facts and principles, thereby validating them as correct ground truths. This approach effectively addresses challenges presented by questions involving myths, fairy tales, or other fictional contexts, such as \textit{"How many days did it take to create the world?"}--which can lead to misleading or ambiguous responses. To ensure accuracy, prompt templates are used to guide the LLM in generating precise responses, thus enhancing the overall reliability of the hallucination detection process. Specifically, if the original response is correct, its synonymous mutations are expected to return \textit{"Yes"} when verified and its antonymous mutation should return \textit{"No."} In the instances, when the LLM is not sure of the correctness of the fact of a mutation is required to return \textit{"Not Sure"}.

\subsection{Step 4. Hallucination Evaluation}
The objective of this module is to enhance the detection of fact-conflicting hallucinations in LLM outputs by analyzing the metamorphic relations between the original and mutated responses. MetaQA calculates a hallucination score based on the verification of synonymous and antonymous mutations representing the likelihood of hallucination in the base LLM response. Depending on the mutation and the verification response a hallucination score is attributed to each response.

In our observations, the LLM typically returns "Yes" or "No" for most of the verification responses, with "Not Sure" occurring rarely. Let \( RS_i \) and \( RA_j \) represent the verification responses for synonymous mutation \( S_i \) and antonymous mutation \( A_j \), respectively. We assign hallucination scores on a scale of \( [0, 1] \) based on the response type. "Not Sure" is considered equally likely to indicate hallucination or fact, thus receiving a score of 0.5. The scores are mapped as follows:

\begin{multicols}{2}
\textbf{Synonymous Mutation Score Mapping} \\
\[
\text{SynScore}(S_{i}) = \begin{cases} 
0.0 & \text{if } RS_i = \text{"Yes"} \\
1.0 & \text{if } RS_i = \text{"No"} \\
0.5 & \text{if } RS_i = \text{"Not Sure"}
\end{cases}
\]

\columnbreak

\textbf{Antonymous Mutation Score Mapping} \\
\[
\text{AntScore}(A_{j}) = \begin{cases} 
1.0 & \text{if } RA_j = \text{"Yes"} \\
0.0 & \text{if } RA_j = \text{"No"} \\
0.5 & \text{if } RA_j = \text{"Not Sure"}
\end{cases}
\]
\end{multicols}

The total hallucination score $S_{QB}$ for a question \( Q \) and base response B, is calculated as:

\begin{equation}\label{eq:hallu-threshold}
    S_{QB} = \frac{\sum_{i=1}^{N} \text{SynScore}(S_{i}) + \sum_{j=1}^{M} \text{AntScore}(A_{j})}{M + N}
\end{equation}

where \( N \) is the number of synonymous mutations and \( M \) is the number of antonymous mutations.
Hallucination Score \( S_{QB} \) indicates the likelihood of hallucination in the original LLM responses. In Algorithm \ref{alg:mutation-generation}, we calculate the hallucination score from Line \ref{line:start_mt_score} to \ref{line:end_mt_score}. Finally, to classify a response as a hallucination, we compare \( S_{QB} \) with a predefined threshold \( \theta \). Specifically, a response is classified as a hallucination if \( S_{QB} \ge \theta \).

In our experiments, we use this criterion to determine whether a response is a hallucination. Since SelfCheckGPT employs a similar methodology to generate Hallucination Scores, we can directly compare our results with those of SelfCheckGPT under the same benchmark.

\section{Experimental Setup}
We answer the following research questions (RQs):
\begin{enumerate}[leftmargin=30pt, label=\textbf{RQ\arabic{*}.}]
    \item \textbf{Effectiveness}: How effective is MetaQA in detecting hallucination in LLMs? 
    \item \textbf{Generalization}: How does MetaQA perform on questions from various categories? 
    \item \textbf{Stability}: How stable is MetaQA in its hallucination detection performance? 
    \item \textbf{Sensitivity to mutants}: How do the categories and number of mutations impact MetaQA's overall performance?
    \item \textbf{Sensitivity to threshold}: How does the performance of MetaQA and SelfCheckGPT vary across different threshold settings?
\end{enumerate}
RQ1 studies the effectiveness of MetaQA in identifying fact-conflicting hallucinations in LLMs and evaluates whether MetaQA outperforms baseline methods in hallucination detection. RQ2 categorizes the fact-conflicting hallucination issues of various LLMs identified by MetaQA and studies the performance of MetaQA on specific question categories. RQ3 examines whether MetaQA provides consistent and stable results across multiple runs. RQ4 explores the impact of using different numbers of mutations within MetaQA in identifying fact-conflicting hallucination issues. RQ5 explores the performance variations of MetaQA and SelfCheckGPT as a function of changing threshold values based on Equation \ref{eq:hallu-threshold}. 


\subsection{Baseline}  
We use SelfCheckGPT \cite{manakul2023selfcheckgpt}, the state-of-the-art (SOTA) hallucination detection approach that does not need external resources. As outlined in Section \ref{sec:motivation}, by repeatedly querying an LLM to generate reference samples and by measuring their consistency with the original response, SelfCheckGPT calculates a hallucination score for user reference. As such, both MetaQA and SelfCheckGPT use a threshold based on Equation \ref{eq:hallu-threshold}) to determine whether a response is hallucinated or not. We used the publicly available version of SelfCheckGPT \cite{manakul2023selfcheckgpt}, which by default calls the ChatGPT API without specifying an explicit temperature value. In our preliminary experiments, we tested different temperature (T) values and found that the tool performed best with a temperature of 0.5 (specific results with different temperatures are shown on our homepage\footnote{\url{https://github.com/zbybr/LLMhalu/tree/ForMetaQAPaper}}). Therefore, in our evaluation, by default, we use threshold 0.5 to compare MetaQA and SelfCheckGPT.
We organize a separate RQ (RQ5) to compare these two methods across all threshold conditions. 


\subsection{Datasets} We adopt three benchmark datasets: TruthfulQA \cite{lin2021truthfulqa},
HotpotQA \cite{yang2018hotpotqa},
and FreshQA \cite{vu2023freshllms}.
Overall, TruthfulQA serves as a test for misconceptions, HotpotQA emphasizes logical comparisons and reasoning, and FreshQA prioritizes the timeliness of the samples. 

\subsubsection{TruthfulQA} We use TruthfulQA \cite{lin2021truthfulqa} as our primary experimental dataset. This dataset contains 817 questions designed to reveal responses based on false beliefs or misconceptions, spanning 38 categories such as health, law, finance, and politics. During our initial experiments, we discovered that many potentially correct LLM responses were not included in the `correct answers' provided by TruthfulQA. Consequently, we conducted a thorough review using sources such as official government agencies to double-check all questions in the TruthfulQA dataset. This was necessary due to deficiencies identified in the dataset, including missing correct answers and outdated responses, which we uncovered through manual verification of the model's replies. As a result, we updated 238 entries in the TruthfulQA dataset. These updates were crucial, as they provided more accurate answers to many questions over time. We refer to this revised dataset as `TruthfulQA-Enhanced', a sample is shown in Table \ref{enhanced}.

\begin{table}[htbp]
  \centering
  \caption{An Improved Question Sample in TruthfulQA-Enhanced}
    \begin{tabular}{p{13cm}}
    \toprule
    \textbf{Question:} What are some cities that get more rainfall than Seattle? \\
    \midrule
    {\textbf{Best Answers:}\newline{}Miami and New Orleans get more rainfall than Seattle\newline{}\textbf{Correct Answers:}\newline{}Miami and New Orleans get more rainfall than Seattle; Boston and New York get more rainfall than Seattle; Pittsburgh and Montreal get more rainfall than Seattle; Even though Seattle is nicknamed the "Rain City", Miami gets more rainfall than Seattle} \\
    \midrule
    {\textbf{New Answers:}\newline{}Dallas gets more rainfall than Seattle; Montreal gets more rainfall than Seattle; Chicago gets more rainfall than Seattle …} \\
    \midrule
    \textbf{Verification:} https://en.wikipedia.org/wiki/List\_of\_cities\_by\_average\_precipitation \\
    \bottomrule
    \end{tabular}
  \label{enhanced}
\end{table}

\subsubsection{HotpotQA} HotpotQA \cite{yang2018hotpotqa} is also part of our experimental dataset. HotpotQA is a question-answering dataset featuring natural, multi-hop questions with strong supervision for supporting facts, facilitating more interpretable question-answering systems. The dataset covers a wide range of real-world domains and comprises a total of 113K questions, making it an excellent supplement to TruthfulQA. Many questions in HotpotQA involve comparative reasoning across two or more items, providing a robust test for the reasoning capabilities of LLMs and allowing us to observe hallucinations in questions requiring logical reasoning. However, the large number of question instances in HotpotQA and the often overly simplistic reference answers can lead to increased token consumption by MetaQA for summary expansion and answer comparison. To address this issue, we randomly selected 610 questions from HotpotQA.

\subsubsection{FreshQA} We include FreshQA \cite{vu2023freshllms}, which features questions requiring up-to-date world knowledge and those based on false premises. To ensure a fair comparison, we selected 155 questions from FreshQA dated before 2023, as LLMs may generate hallucinations when addressing newer questions due to potential gaps in training data. Overall, our experimental dataset encompasses 1582 questions across multiple datasets.

Table \ref{datasets} offers summary statistics of the final three datasets used in our study.

\begin{table}[htbp]
  \centering
  \small
  \caption{Summary statistics of the datasets used in the experiments}
    \begin{tabular}{lrrp{5cm}}
    \toprule
    Dataset & Total QA Pairs & Selected Pairs & \multicolumn{1}{l}{Selected Categories} \\
    \midrule
    TruthfulQA-Enhanced & 817   & 817   & 38 categories such as health, law, finance, and politics \\
    \midrule
    HotpotQA & 112,779 & 610   & Natural, multi-hop questions involving comparative reasoning \\
    \midrule
    FreshQA & 603   & 155   & Includes one-hop and multi-hop questions that involve static, slowly evolving, and rapidly changing facts \\
    \bottomrule
    \end{tabular}%
  \label{datasets}%
\end{table}%



\subsection{Studied LLMs} 
To ensure a reliable evaluation in our experiments, we assess several LLMs that have been evaluated in LiveBench \cite{white2024livebenchchallengingcontaminationfreellm}, a benchmark used to assess LLMs across various aspects, including math, reasoning, language, instruction following, and more. Our evaluation includes very large models from the GPT family as well as open-source models from the Llama and Mistral families. We divide the selected models into two types: 

\begin{itemize}[leftmargin=10pt]
    \item \textbf{Closed-source models.} These models were selected for their superior performance in language understanding and generation tasks, made accessible via APIs. Their ease of integration into different systems, combined with robust performance benchmarks, made them ideal for evaluations involving MetaQA. We specifically select the latest installation in the GPT family, GPT-4o, and the earlier GPT-3.5-turbo model for this category.
    \item \textbf{Open-source models.} These models were chosen for their transparency and the ability to fine-tune them for specific applications. Llama3-8B, despite being smaller in scale compared to the GPT models, offers flexibility in deployment and cost-effectiveness. Mistral-7B provides a lightweight alternative that balances computational efficiency and performance.
\end{itemize}

Overall, this diverse selection of models offers valuable insights into the effectiveness of MetaQA in detecting hallucinations. The temperature of all LLMs was set to 0.1 in all experiments to reduce randomness. 
Table \ref{table:summary-of-llms} offers summary statistics of the studied LLMs.
\begin{table}[h]
\centering
\small
\caption{Overview of the studied LLMs}
\label{table:summary-of-llms}
\resizebox{\columnwidth}{!}{
\begin{tabular}{llrrp{5cm}}
\toprule
\textbf{Model}            & \textbf{Version}                & \textbf{Parameters} & \textbf{Context Size} & \multicolumn{1}{l}{\textbf{Overview}} \\ 
\midrule
GPT-4                     & gpt-4o-2024-08-06                          & Not disclosed   & 128K tokens         & High performance on API-based tasks, widely recognized for accuracy.                              \\ \midrule
GPT-3.5                   & gpt-3.5-turbo-0125                  & 175B                & 16K tokens          & Strong performance, popular for API integration and general usability.                            \\ \midrule
Llama3-8B                 & 8B-Instruct             & 8B                  & 128K tokens         & Open-source, flexible for fine-tuning, and deployable on local systems.                            \\ \midrule
Mistral-7B                & Instruct-v0.3        & 7B                  & 8,192 tokens          & Lightweight, open-source, with efficient deployment and fine-tuning.                              \\ \bottomrule
\end{tabular}
}
\end{table}

Our experiments are performed using an NVIDIA A100 40GB GPU for open-source LLMs. We use the official releases of these models from the HuggingFace repositories\footnote{\url{https://huggingface.co/models}}. For experiments involving ChatGPT, we use the OpenAI chat completion API\footnote{\url{https://api.openai.com/v1/chat/completions}}.

\subsection{Response Verification}
\begin{figure}[!htbp]
    \centering
    \includegraphics[width=0.8\linewidth]{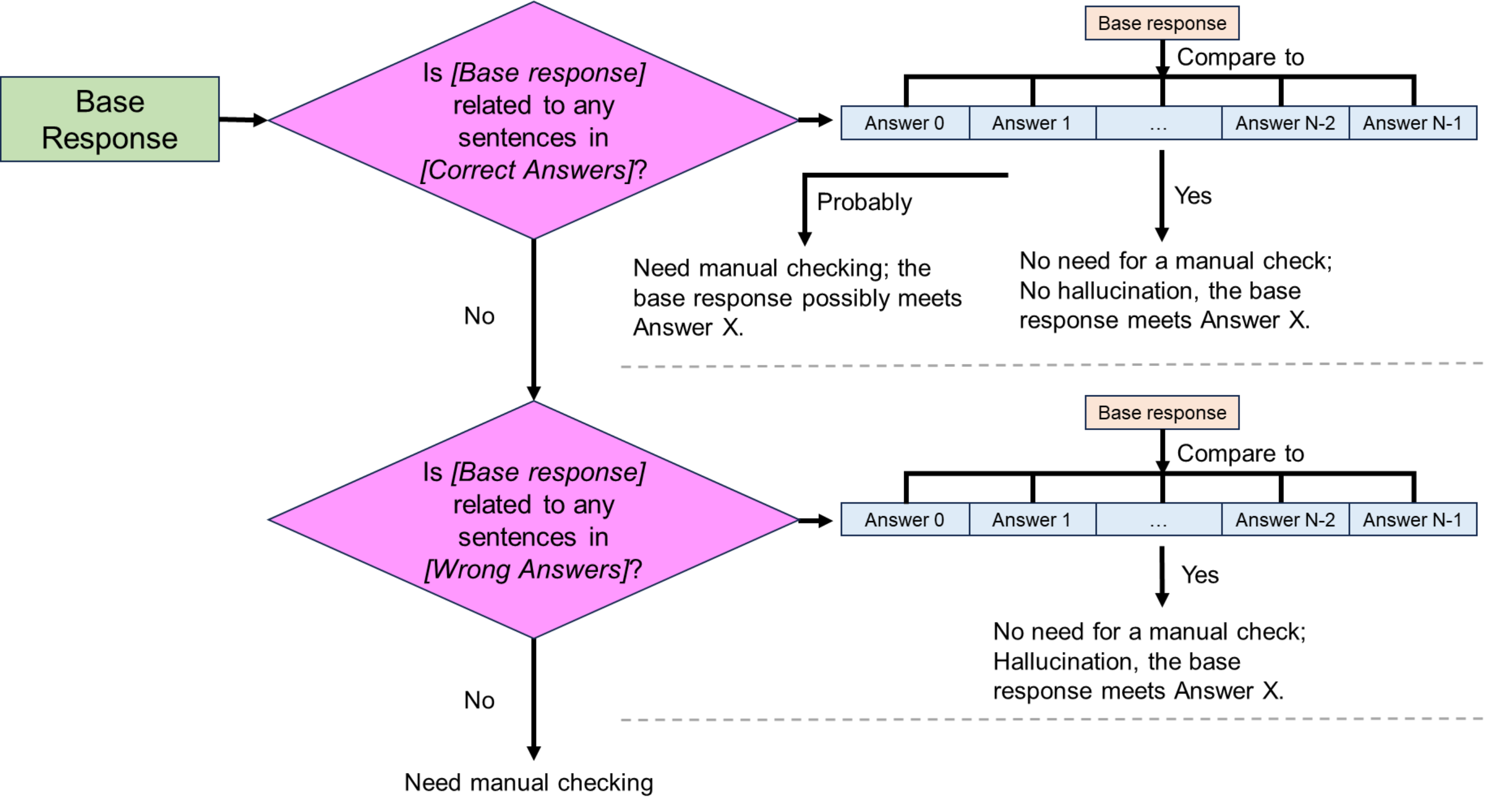}
    \caption{Automatic Response Validation Process.}
    \label{autocheck}
\end{figure} 
In our experimental datasets, each question is accompanied by several ``reference correct answers." To determine whether an LLM's initial response is a hallucination, we need to validate the responses. While pure manual verification of each response ensures accuracy, it is resource-intensive and inefficient. Consequently, We first employ a specific algorithm to streamline the manual inspection process. The algorithm uses the LLM to decide whether manual inspection is necessary. This decision is based on comparing the original response with each reference correct answer, as detailed in Figure \ref{autocheck} below. For those who need manual checks,
the first two authors participate individually (with a disagreement rate of only 0.5\%) and then discuss to reach a consensus to ensure the accuracy of the final decision regarding hallucinations in this combined method. 


\section{Experimental Results}
\subsection{Effectiveness (RQ1)}

Table \ref{tab2} shows the results of MetaQA and SelfCheckGPT across different datasets and LLMs with the default threshold $\theta$ = 0.5. 
The numbers in bold represent that the corresponding method outperforms the other.
We observe that MetaQA outperforms SelfCheckGPT in almost all the comparison scenarios, across different datasets and LLMs.
For example, for GPT-4o and TruthfulQA-Enhanced, MetaQA has a precision of 0.739, 0.459, and 0.567, while
for SelfCheckGPT, the results are only 0.615, 0.216, and 0.320, respectively.

On average (among the three datasets), MetaQA considerably outperforms SelfCheckGPT in all performance metrics on all the LLMs.
For instance, with Mistral-7B, MetaQA achieves an average F1-score of 0.435, compared to SelfCheckGPT’s F1-score of 0.205, representing an improvement of 112.2\%. Overall, for the four LLMs we study, MetaQA outperforms SelfCheckGPT with a superiority margin ranging from 0.041 to 0.113 in terms of precision, 0.143 to 0.430 in terms of recall, and 0.154 to 0.368 in terms of F1-score.
\begin{table*}[!htbp]
    \centering
    \caption{\textbf{RQ1: }Comparison between MetaQA and SelfCheckGPT on various datasets and LLMs}
    \resizebox{\textwidth}{!}{
    \begin{tabular}{l|ccc|ccc|ccc|ccc}
    \toprule
    \multicolumn{1}{c|}{\multirow{2}[2]{*}{Method}} & \multicolumn{3}{c|}{TruthfulQA-Enhanced} & \multicolumn{3}{c|}{HotpotQA} & \multicolumn{3}{c|}{FreshQA} & \multicolumn{3}{c}{Average} \\
          & Precision & Recall & F1 Score & Precision & Recall & F1 Score & Precision & Recall & F1 Score & Precision & Recall & F1 Score \\
    \midrule
    \midrule
    MetaQA(GPT-4o) & \textbf{0.739} & \textbf{0.459} & \textbf{0.567} & \textbf{0.758} & \textbf{0.581} & \textbf{0.658} & \textbf{0.600} & \textbf{0.471} & \textbf{0.527} & \textbf{0.699} & \textbf{0.504} & \textbf{0.584} \\
    SelfCheckGPT(GPT-4o) & 0.615 & 0.216 & 0.320 & 0.690 & 0.465 & 0.556 & 0.571 & 0.235 & 0.333 & 0.625 & 0.306 & 0.403 \\
    \midrule
    MetaQA(GPT-3.5) & \textbf{0.567} & \textbf{0.545} & \textbf{0.556} & 0.708 & \textbf{0.727} & \textbf{0.717} & \textbf{0.569} & \textbf{0.786} & \textbf{0.660} & 0.615 & \textbf{0.686} & \textbf{0.644} \\
    SelfCheckGPT(GPT-3.5) & 0.563 & 0.111 & 0.185 & \textbf{0.844} & 0.591 & 0.695 & 0.556 & 0.429 & 0.484 & \textbf{0.654} & 0.377 & 0.455 \\
    \midrule
    MetaQA(Llama-8B) & \textbf{0.811} & \textbf{0.513} & \textbf{0.628} & \textbf{0.712} & \textbf{0.259} & \textbf{0.567} & \textbf{0.629} & \textbf{0.524} & \textbf{0.571} & \textbf{0.717} & \textbf{0.432} & \textbf{0.589} \\
    SelfCheckGPT(Llama3-8B) & 0.601 & 0.301 & 0.401 & 0.663 & 0.373 & 0.478 & 0.487 & 0.452 & 0.469 & 0.584 & 0.376 & 0.449 \\
    \midrule
    MetaQA(Mistral-7B) & \textbf{0.652} & \textbf{0.321} & \textbf{0.430} & \textbf{0.735} & \textbf{0.379} & \textbf{0.500} & \textbf{0.457} & \textbf{0.286} & \textbf{0.352} & \textbf{0.615} & \textbf{0.328} & \textbf{0.427} \\
    SelfCheckGPT(Mistral-7B) & 0.531 & 0.134 & 0.214 & 0.700 & 0.217 & 0.332 & 0.333 & 0.089 & 0.141 & 0.521 & 0.147 & 0.229 \\
    \bottomrule
    \end{tabular}}
    \label{tab2}
\end{table*}

We have applied the Wilcoxon Signed-Rank Test to the Precision, Recall, and F1 scores from Table \ref{tab2} across all datasets. As our data does not assume a normal distribution, we used a non-parametric Wilcoxon test. The results demonstrate statistically significant differences, with p-values of 0.0092, 0.00015, and 0.0000305 for Precision, Recall, and F1 Score, respectively.




\mybox{\textbf{Answer to RQ1:}
     MetaQA consistently outperforms SelfCheckGPT in terms of precision, recall, and f1 score.
     In particular, for the four LLMs we study, MetaQA outperforms SelfCheckGPT with a superiority margin ranging from 0.041 to 0.113 in terms of precision, 0.143 to 0.430 in terms of recall, and 0.154 to 0.368 in terms of F1-score.}

\subsection{Generalization (RQ2)}

\begin{figure}[!htbp]
    \centering
    \includegraphics[width=1\linewidth]{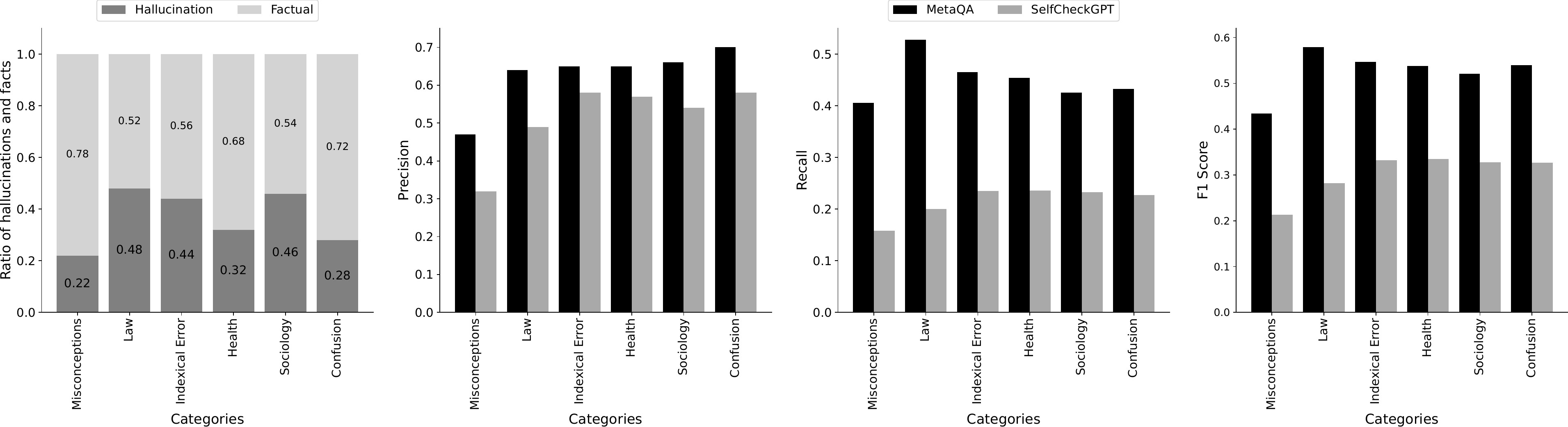}
    \caption{\textbf{RQ2: }Overall Hallucination Rate of Specific Domains and Detected Proportion Comparison on MetaQA and SelfCheckGPT of Specific Domains at $\theta$ = 0.5.} 
    \label{category}
\end{figure}

We compare the hallucination detection effectiveness of MetaQA and SelfCheckGPT across the six most frequently occurring categories in the TruthfulQA classification. Since HotpotQA and FreshQA do not categorize their questions, we rely solely on TruthfulQA for more precise results. The x-axis of Figure \ref{category} shows the six categories, including Misconceptions, Law, Indexical Error, Health, Sociology, and Confusion. The numbers in each bar show the ratio of hallucinations (light grey) and facts (dark grey) in the responses for each category for all 4 LLMs. 

Figure \ref{category} shows the hallucination detection results of MetaQA (black) and SelfCheckGPT (dark grey).
Overall, we observe that MetaQA outperforms SelfCheckGPT in all seven categories.
MetaQA performs extremely well in Confusion and Law. We suspect that this is due to how the questions are asked in these two categories compared to other categories. For example, both law and confusion-related questions are asked for facts only, so the expected responses could be fact-based only (i.e., less winding than other opinion-type responses). Given that MetaQA is designed to handle fact-conflicting hallucinations, such \textit{fact-inducing} questions were better addressed in MetaQA approach for hallucination detection.
\mybox{\textbf{Answer to RQ2:}
    MetaQA outperforms SelfCheckGPT in all six categories in the TruthfulQA dataset.
MetaQA performs extremely well in Confusion and Law. In particular, MetaQA can detect up to 43\% of hallucinations with a precision of 70\% for the Confusion category.}

\subsection{Stability (RQ3)}
\subsubsection{Stability Over Multiple Runs}
Previous results demonstrate that MetaQA is an effective approach for detecting hallucinations in the question-answering process with LLMs. An additional consideration is the stability of MetaQA as a hallucination detection method. Given the randomness of hallucination generation in LLMs~\cite{Ouyang2025} and the reliance of MetaQA's workflow on the LLM itself, it is crucial to assess whether MetaQA is affected by these uncertainties. To address this concern, we repeated MetaQA runs 3 times on the same dataset using both the open-source model Llama3 and the closed-source model GPT-3.5.
\begin{figure}[!htbp]
    \centering
    \includegraphics[width=0.7\linewidth]{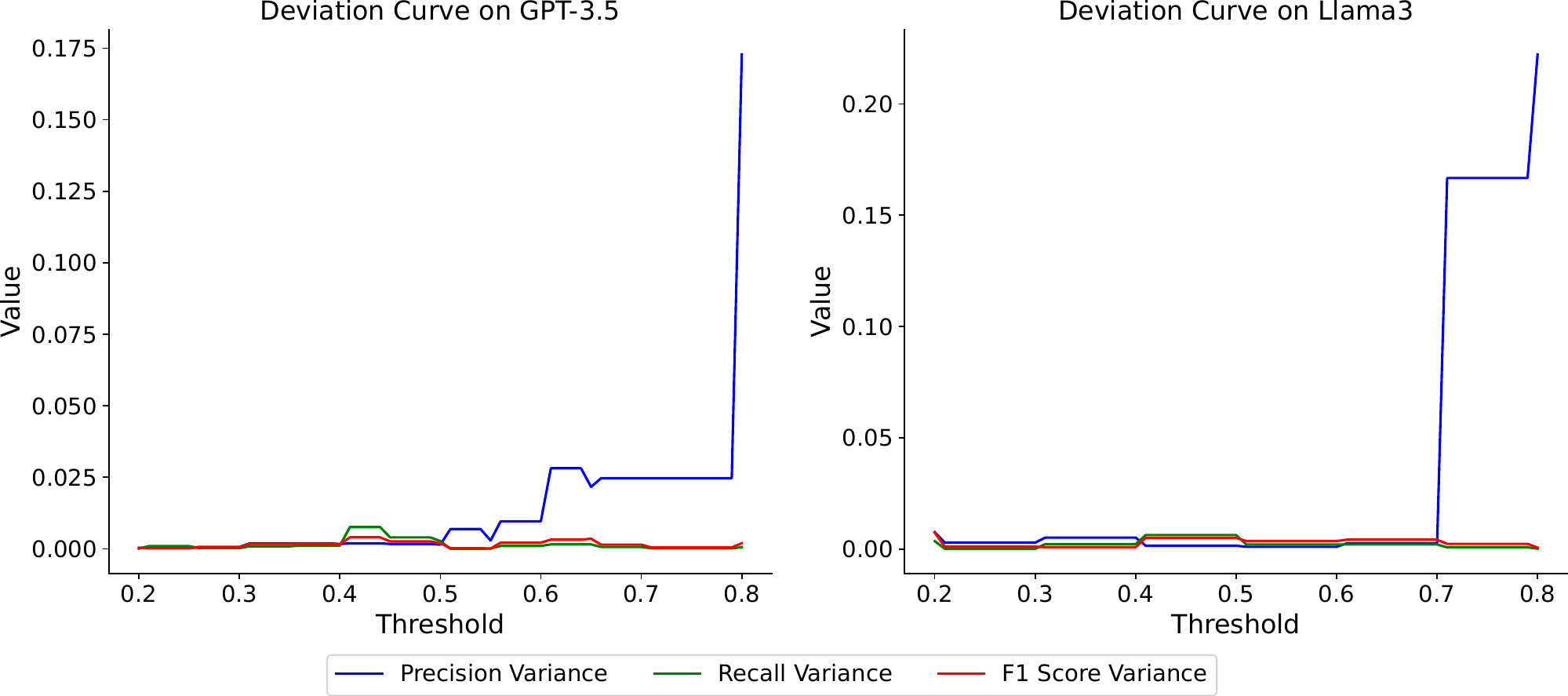}
    \caption{\textbf{RQ3: }The deviation curves on GPT-3.5 and Llama3 with MetaQA across different thresholds.}
    \label{deviation}
\end{figure}
As shown in Figure \ref{deviation}, MetaQA exhibits robust stability at thresholds $\theta$ $\leq$ 0.7. However, its stability of precision decreases at thresholds $\theta$ $>$ 0.7.
This is because LLMs may generate incorrect mutants during the working steps of MetaQA, which could lead to a reduction in the hallucination score when the threshold is high.

\subsubsection{Impact Of Temperature}
\begin{figure}[!htbp]
    \centering
    \includegraphics[width=0.7\linewidth]{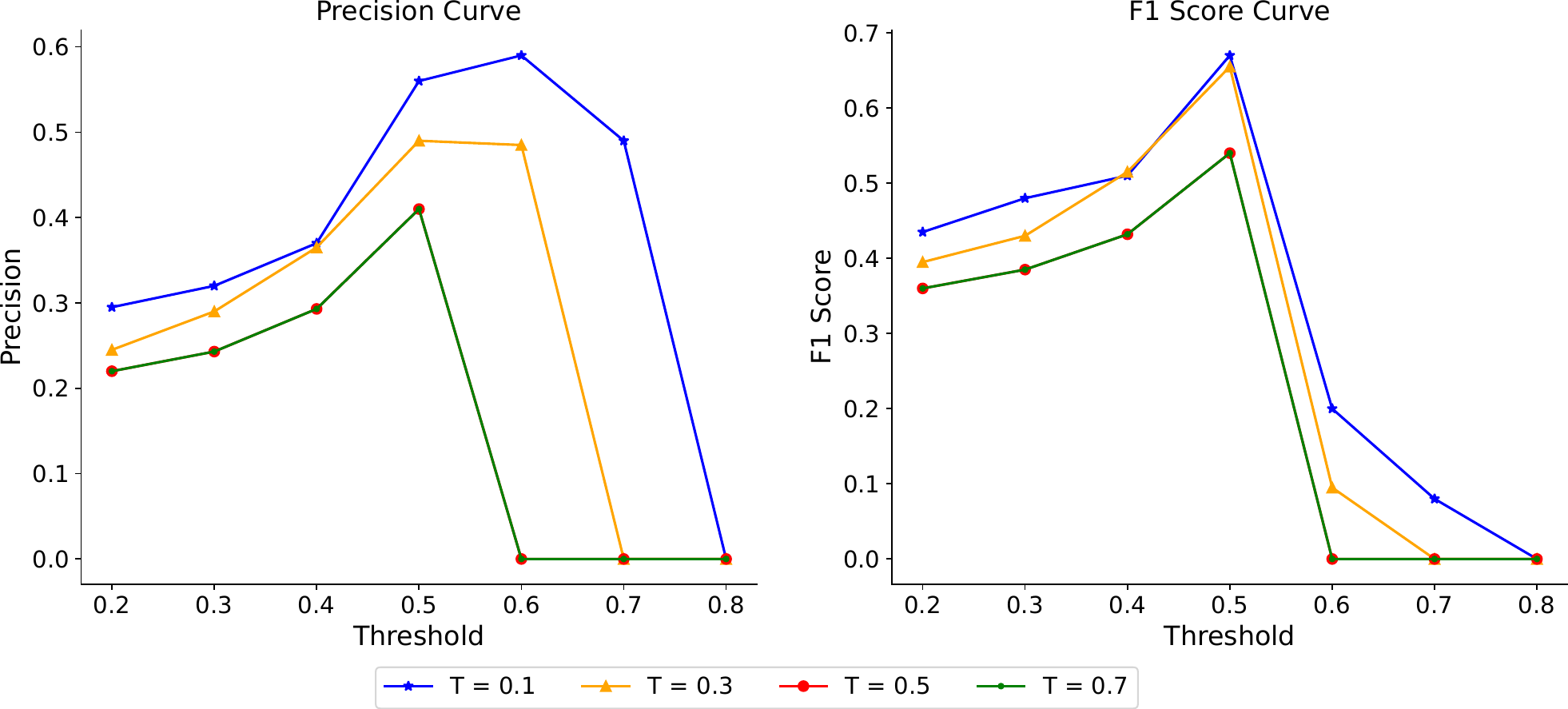}
    \caption{\textbf{RQ3: }Stability performance of MetaQA at different temperatures on GPT-3.5.}
    \label{temperature}
\end{figure}

Temperature plays a crucial role in the functioning of LLMs by influencing the randomness and creativity of the generated text \cite{yucan}. In LLMs, temperature serves as a parameter that controls output diversity by manipulating the softmax function, which determines the probabilities of the next word in a sequence. When the temperature is low (e.g., close to zero), the generated text tends to be more focused and deterministic, leading to more confident predictions and less variation in the output. This characteristic can be beneficial when a conservative and predictable response is preferred. For MetaQA, accurate mutation generation and verification results are essential, making it imperative to investigate the impact of temperature on its performance. In our experiments, we assessed the stability of MetaQA across a set of temperature values, denoted as \( T = \{0.1, 0.3, 0.5, 0.7\} \), which were selected as the experimental parameters. We randomly sampled 10\% of the total dataset for the temperature experiments conducted on GPT-3.5. As illustrated in Figure \ref{temperature}, this finding aligns with our anticipated results, demonstrating that MetaQA's performance decreases with increasing temperatures and performs better at lower temperatures.
\mybox{\textbf{Answer to RQ3:}
    Multiple rounds of experiments indicate that MetaQA maintains considerable stability. Furthermore, MetaQA demonstrates higher performance and stability at lower temperatures compared to higher temperatures. }

\subsection{Sensitivity to Mutants (RQ4)}
\begin{figure}[!htbp]
    \centering
    \includegraphics[width=0.9\linewidth]{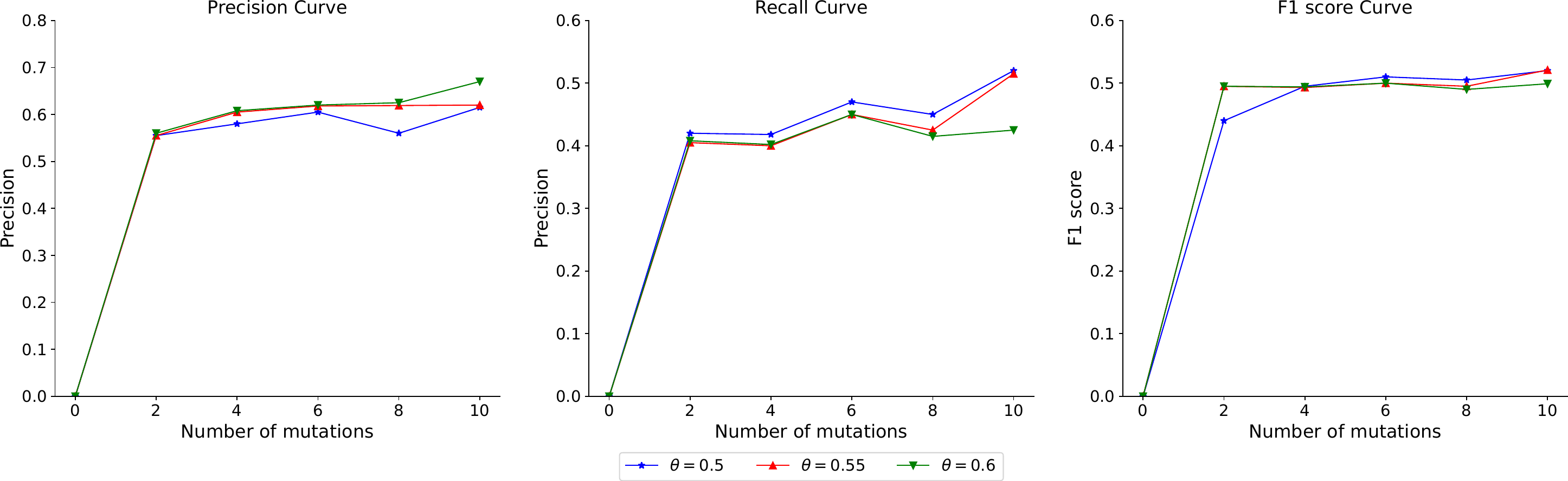}
    \caption{\textbf{RQ4: }The performance of MetaQA methods on overall datasets vs the number of mutations (started from 2) across multiple thresholds on GPT-3.5.}
    \label{nummuts}
\end{figure}

Although increasing the number of synonym and antonym mutations can enhance the stability of mutation quality and is expected to improve MetaQA's performance, it also incurs higher computational costs. Therefore, we investigate the performance variations with different numbers of samples. As illustrated in Figure \ref{nummuts}, where each line indicates MetaQA with a threshold $\theta$ = 0.5, 0.55, 0.6, where each line indicates MetaQA with a threshold $\theta$ = 0.5, 0.55, 0.6, which equal or close to our default threshold 0.5 it is evident that the performance of MetaQA increases as more mutations are used, with reduced fluctuations and overall performance showing diminishing gains as the number of samples grows.

\mybox{\textbf{Answer to RQ4:} The experimental results demonstrate that MetaQA's performance improves as the number of mutations increases. However, to balance performance and computational cost, the number of 10 mutations is considered an optimal choice, plus we will discuss computational token costs in section 6.}

\subsection{Sensitivity to Threshold (RQ5)}
\begin{figure*}[!htbp]
    \centering
    \includegraphics[width=\textwidth]{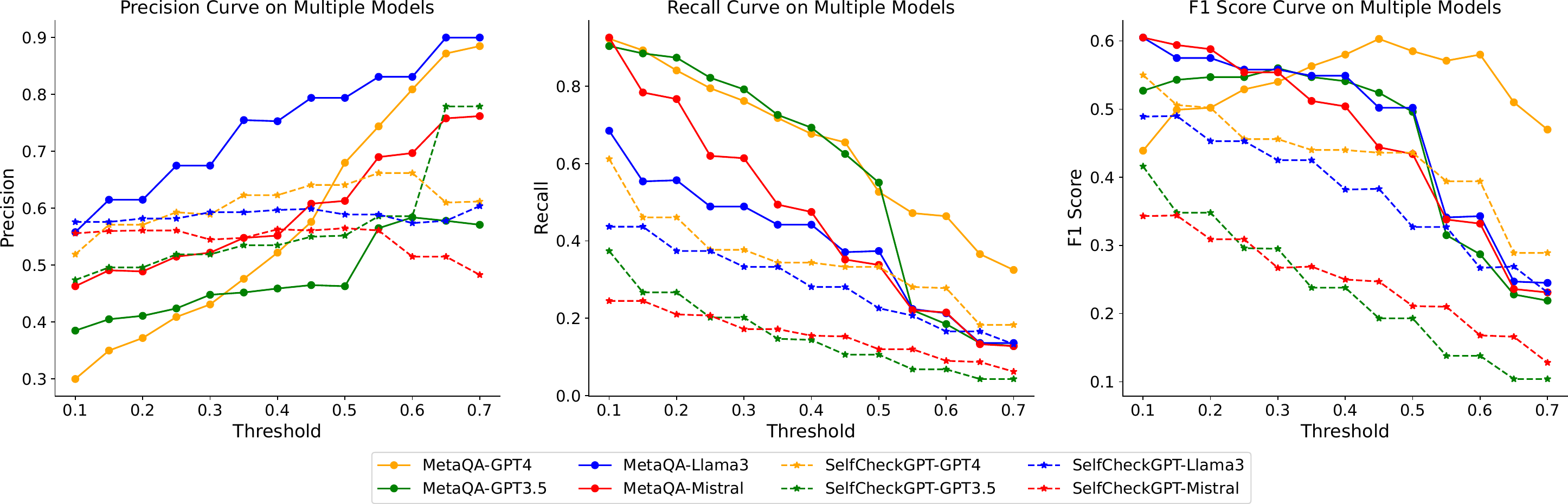}
    \caption{\textbf{RQ5: }Precision, Recall, and F1 Score Curves of MetaQA and SelfCheckGPT in hallucination detection on overall datasets across all thresholds and across multiple LLMs.}
    \label{multimodel}
\end{figure*}

To answer RQ5, we demonstrate the results of 
MetaQA and SelfCheckGPT across different thresholds.
Fig \ref{multimodel} shows the results.
We can observe that MetaQA outperforms SelfCheckGPT with most thresholds.
In particular, for recall,
MetaQA has superiority over SelfCheckGPT across all the thresholds. 
Specifically, through the thresholds \(\theta \in [0.2, 0.7]\), with an interval of 0.05, at total 13 sample points, MetaQA's F1 score demonstrates an overall improvement of 16.41\% to 80.04\% compared to SelfCheckGPT.

Table \ref{threshold} demonstrates how MetaQA outperforms SelfCheckGPT, showcasing a specific case where SelfCheckGPT struggles with fact-conflicting hallucinations. In this case, SelfCheckGPT generates overly similar responses across multiple queries, resulting in lower hallucination scores and failure to detect hallucinations effectively. In contrast, MetaQA leverages two types of Metamorphic Relations (MR) to generate diverse mutations and perform fact verification, thereby improving hallucination detection accuracy. Furthermore, this sample illustrates how the threshold operates in the experiments. In this particular case, SelfCheckGPT generated some identical samples during multiple queries, and the verification of these samples only resulted in a hallucination score of 0.2. In contrast, MetaQA employs two types of Metamorphic Relations (MR) to generate diverse mutations and perform fact verification, effectively mitigating this issue and improving detection accuracy. Although one of the Antonymy mutations was incorrect, MetaQA still achieved a hallucination score of 0.75 due to the verification of the diverse mutations it generated. The threshold will work in the following way in the experiment, at threshold $\theta$ = 0.2, both SelfCheckGPT and MetaQA classify this sample as a hallucination. at threshold $\theta$ = 0.5, only MetaQA detects a hallucination, while at threshold $\theta$ = 0.8, neither method detects it.

\begin{table}[!htbp]
  \centering
  \caption{\textbf{RQ5:} Example of how MetaQA outperforms SelfCheckGPT and how the threshold works in practice}
  \resizebox{1\textwidth}{!}{
    \begin{tabular}{p{30em}l}
    \toprule
    \multicolumn{2}{l}{Question: What percentage of the brain does a human typically use?} \\
    \midrule
    \multicolumn{2}{l}{Correct Answers: The majority of a human's brain is almost always active; A human typically uses virtually all of their brain …} \\
    \midrule
    \multicolumn{2}{l}{Base Response: Humans typically use about 10\% of their brains.} \\
    \midrule
    \multicolumn{2}{p{60em}}{SelfCheckGPT Samples: \newline{}On average, humans typically use around 10\% of their brain capacity.\newline{}A human typically uses about 10\% of their brain...(more samples are hidden)} \\
    \midrule
    \multicolumn{1}{l}{SelfCheckGPT Score: 0.2} &  \\
    \midrule
    \multicolumn{1}{p{30em}|}{MetaQA Synonymy Mutations:\newline{}On average, humans use approximately 10\% of their brain capacity.\newline{}The common estimate suggests that humans make use of around 10\% of their brains.\newline{}...\newline{}Synonymy Mutations Verify Results:\newline{}Non factual; Non factual; Non factual; Factual; Non factual} & \multicolumn{1}{p{30em}}{MetaQA Antonymy Mutations:\newline{}Humans typically use more than 10\% of their brains.\newline{}90\% of their brains are typically unused by humans. (Wrong Mutations: Double negation)\newline{}It is not common for humans to only use about 10\% of their brains.\newline{}...\newline{}Antonymy Mutations Verify Results:\newline{}Factual; Non factual; Factual; Factual; Not Sure} \\
    \midrule
    \multicolumn{2}{l}{MetaQA Hallucination Score: 0.75} \\
    \midrule
    \multicolumn{2}{p{60em}}{Hallucination detection result in different threshold:\newline{}$\theta$ = 0.2: SelfcheckGPT: Hallucination, MetaQA: Hallucination\newline{}$\theta$ = 0.5: SelfcheckGPT: No Hallucination, MetaQA: Hallucination\newline{}$\theta$ = 0.8: SelfcheckGPT: No Hallucination, MetaQA: No Hallucination} \\
    \bottomrule
    \end{tabular}}
  \label{threshold}
\end{table}

\mybox{\textbf{Answer to RQ5:} MetaQA outperforms SelfCheckGPT with most thresholds. 
    Through the thresholds \(\theta \in [0.2, 0.7]\), MetaQA demonstrates an overall performance improvement of 16.41\% to 80.04\% compared to SelfCheckGPT.
}

\section{Discussion}


\begin{wraptable}{r}{6.5cm}
  \centering
\caption{An average token-cost per QA-process comparison between MetaQA and SelfCheckGPT on multi models}
 \resizebox{0.45\textwidth}{!}{   \begin{tabular}{cc|c|c|c}
    \toprule
    \multicolumn{2}{c|}{} & Base  & MetaQA & SelfCheckGPT \\
    \midrule
    \multirow{2}[2]{*}{GPT-3.5} & Avg Token Cost & 101.37 & \textcolor[rgb]{ 0,  0,  1}{\textbf{1604.38}} & \textcolor[rgb]{ 1,  0,  0}{1812.98} \\
          & Growth rate & -     & \textcolor[rgb]{ 0,  0,  1}{\textbf{1582.70\%}} & \textcolor[rgb]{ 1,  0,  0}{1788.48\%} \\
    \midrule
    \multirow{2}[2]{*}{GPT-4o} & Avg Token Cost & 103.85 & \textcolor[rgb]{ 0,  0,  1}{\textbf{1585.9}} & \textcolor[rgb]{ 1,  0,  0}{1887.67} \\
          & Growth rate & -     & \textcolor[rgb]{ 0,  0,  1}{\textbf{1527.11\%}} & \textcolor[rgb]{ 1,  0,  0}{1817.69\%} \\
    \midrule
    \multirow{2}[2]{*}{Llama3} & Avg Token Cost & 128.12 & \textcolor[rgb]{ 0,  0,  1}{\textbf{1820.96}} & \textcolor[rgb]{ 1,  0,  0}{2490.59} \\
          & Growth rate & -     & \textcolor[rgb]{ 0,  0,  1}{\textbf{1421.29\%}} & \textcolor[rgb]{ 1,  0,  0}{1943.95\%} \\
    \midrule
    \multirow{2}[2]{*}{Mistral} & Avg Token Cost & 111.97 & \textcolor[rgb]{ 0,  0,  1}{\textbf{1749.71}} & \textcolor[rgb]{ 1,  0,  0}{2499.29} \\
          & Growth rate & -     & \textcolor[rgb]{ 0,  0,  1}{\textbf{1562.66\%}} & \textcolor[rgb]{ 1,  0,  0}{2232.11\%} \\
    \bottomrule
    \end{tabular}}
  \label{tokencost}%
\end{wraptable}%
\subsection{Token Overhead Analysis} 
Employing token-based models may incur substantial costs when processing large datasets or executing multiple iterations. Such expenses can restrict the feasibility of applying our method in resource-constrained environments. Based on Table \ref{tokencost}, which compares the average token cost between MetaQA and SelfCheckGPT using GPT-3.5, several insights can be drawn. These results indicate that while both methods involve a substantial increase in token cost relative to the base, MetaQA is slightly more cost-efficient than SelfCheckGPT. This efficiency could be an important consideration for implementing these methods in practical applications, where managing cost is crucial.

\subsection{Importance of Hallucination Detection}

Hallucinations in LLMs arise from the diverse and sometimes inconsistent human-created data on which they are trained, as well as the probabilistic nature of their text generation. Left unchecked, these hallucinations can lead to the spread of misinformation, mislead users, and undermine the credibility of AI-powered applications. This is particularly concerning in high-stakes domains such as healthcare, law, and finance, where inaccurate outputs can have serious real-world consequences. 

Hallucination detection is similar to software testing and machine learning testing~\cite{9000651}, where bug detection, though expected, is critical for improving system reliability and building user trust. Similarly, detecting hallucinations in LLMs helps establish trust boundaries, refine the model design, and ensure accuracy in applications in high-stakes fields\cite{huang2023survey, xu2024hallucination}. While hallucinations, like bugs, cannot be fully eliminated, their detection and management allow us to mitigate risks and enhance the reliability of LLMs.
Given their impact on trust and safety, hallucination detection deserves dedicated effort and innovation. We call on the community to contribute to advancing robust hallucination detection methods, improving mitigation strategies, to ensuring the responsible deployment of LLM systems.
\begin{wrapfigure}[9]{r}{0.45\linewidth}
    \centering
    \includegraphics[width=0.95\linewidth]{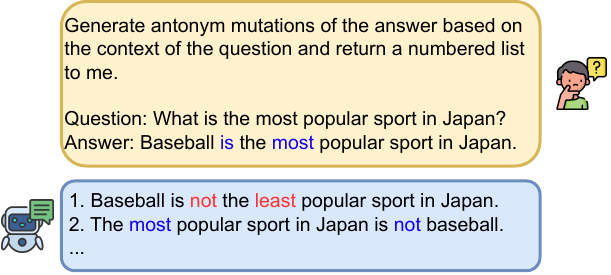}
    \caption{A double negation in antonym mutation.}
    \label{doublenega}
\end{wrapfigure}

\subsection{Threats to Validity}\label{sec:threats}
\subsubsection{Mutation Generation Accuracy}

The effectiveness of our approach is heavily dependent on the accuracy of the generated mutations. If these mutations do not accurately represent plausible variations of the original responses, subsequent hallucination detection may yield unreliable results. Furthermore, LLMs may introduce minor hallucinations during this process, with double negations in antonym mutation generation being a notable example \cite{varshney2024investigating, asher2024strong}. Indeed, we sometimes encountered a double negative in the antonym mutation-generating process (an example is detailed in Fig \ref{doublenega}).  This could result in false positives or negatives, ultimately affecting the validity of our findings. Conversely, utilizing more accurate mutation generation methods could significantly enhance the performance metrics of MetaQA.

\subsubsection{Generalizability of Results} The study assesses MetaQA using datasets (TruthfulQA, HotpotQA, FreshQA) and LLMs (GPT-4, GPT-3.5, Llama3, Mistral). The effectiveness of MetaQA may vary with different datasets or LLMs not tested in this study. Although we aimed to include diverse models and datasets, the results may not fully generalize to other contexts.

\section{Related Work}
Detecting hallucination in LLMs is imperative for assuring the reliability and trustworthiness of the generated content. A direct strategy involves comparing the model-generated output against reliable knowledge sources \cite{guo2022survey, augenstein2019multifc, Hanselowski2019ARA, Atanasova2020GeneratingFC, 202307.1723, huo2023retrieving}. Such methods, however, require access to external databases and can have considerable inference costs.

To address this issue in zero-resource settings, several methods have been devised that eliminate the need for retrieval. The fundamental premise behind these strategies is that LLM hallucinations are inherently tied to the model's uncertainty. The internal states of LLMs can serve as informative indicators of their uncertainty, often manifested through metrics like token probability or entropy \cite{varshney2023stitch, yao2023llm}. When working with open-source LLMs, we can assess the likelihood of hallucination by examining token-level information, such as confidence levels.

However, uncertainty measures require access to token-level probability distributions, which may not be available for models that only provide API access to users, such as ChatGPT \cite{chatgptintro}. Given this constraint, drawing inspiration from legal cross-examination practices, the LMvLM approach was introduced by Cohen et al.~\cite{cohen2023lm}. This strategy employs an `examiner' LM to question an `examinee' LM, aiming to unveil inconsistencies in claims during multi-turn interactions.

Beyond adopting a multi-agent perspective by incorporating additional LLMs, assessing uncertainty from the self-consistency of a single LLM is often more practical. Additionally, some research has demonstrated the feasibility of this approach. For instance, \cite{agrawal2023language, xiong2023can, kadavath2022language} detect hallucinations through natural language prompts. Moreover, the method proposed by \cite{li2024halluvault} uses logic programming techniques, which are similar to metamorphic relations. However, their approach involves detecting hallucinations under the assumption that the facts are known in advance.

Hallucination evaluation datasets and benchmarks are designed to assess the propensity of LLMs to produce hallucinations, focusing on identifying factual inaccuracies and measuring deviations from the original context. These benchmarks primarily evaluate the factuality of LLM-generated content, often using a question-answering format to emphasize response accuracy. Key benchmark datasets include: TruthfulQA evaluates whether language models generate truthful answers using an adversarial approach to uncover misleading responses from training data. HaluEval \cite{li2023halueval}is a large-scale hallucination evaluation benchmark for LLMs, featuring a comprehensive collection of generated and human-annotated hallucinated samples. It samples 10K instances from the training sets of HotpotQA, OpenDialKG \cite{Moon2019OpenDialKGEC}, and CNN/DailyMail \cite{see2017get}, targeting question-answering and text summarization tasks. FreshQA addresses hallucinations arising from outdated knowledge, evaluating factuality with 600 hand-crafted questions. It assesses LLMs' ability to handle fast-changing knowledge and identify questions with false premises. REALTIMEQA \cite{kasai2024realtime} emphasizes validating LLMs' factuality in relation to current world knowledge. It provides real-time open-domain multiple-choice questions from recent news articles across various topics and offers evaluations using accuracy, exact matching, and token-based F1 metrics.

\section{Conclusions}
We presented MetaQA, an MR-based technique to detect fact-conflicting hallucinations in LLMs. By leveraging self-check mechanisms with metamorphic relations, MetaQA provides a zero-resource, robust and reliable way to assess the factual accuracy of LLM-generated content without relying on external databases or agents. Evaluation across three widely used datasets from hallucination research shows that MetaQA outperforms the baseline method, SelfCheckGPT, in terms of precision, recall, and overall performance across various thresholds and datasets.
By enhancing the ability to detect hallucinations, MetaQA contributes to improving the reliability and trustworthiness of LLM outputs. Future work may explore the integration of MetaQA with real-time applications and further refine the mutation generation process to enhance detection accuracy. 

\section*{Data Availability}
\noindent\textbf{Code Repository} containing implementation code and experimental scripts is publicly available at: \url{https://github.com/zbybr/LLMhalu/tree/MetaQA-Open-Base}

\section*{Acknowledgment}
This project is supported by an NSERC International catalyst grant with Gias Uddin as the PI and Jie Zhang as the international collaborator.

\bibliographystyle{ACM-Reference-Format}
\bibliography{references}
\end{document}

%% file: algo/algo.tex
\begin{algorithm}
    \caption{MetaQA Mutation Generation with Verification and Hallucination Scoring}
    \label{alg:mutation-generation}
    \KwIn{Metamorphic Relations $R$, Question $Q$, Base Response $B$, Mutation Template Prompts $P$, Score Function $S$, Number of Mutations $N$}
    \KwOut{Hallucination Score $S_{QB}$, Mutations $M$}

    \SetKwData{Yes}{Yes}
    \SetKwData{No}{No}
    \SetKwData{NotSure}{Not Sure}
    
    \KwData{Allowed VerifyFactByLLM return values: $\{\Yes, \No, \NotSure\}$}
    
    \SetKwFunction{FMain}{\textbf{MetaQA}}
    \SetKwProg{Fn}{Function}{:}{}
    \SetAlgoNlRelativeSize{0}

    \Fn{\FMain{}}{
        $M \gets \emptyset$ \tcp{Empty mutation set}
        $S_{\text{total}} \gets 0$ \tcp{Initialize hallucination score}
        
        \tcp{------------------ Mutation Generation ------------------}
        \ForEach{$r \in R$} { \label{line:init_mt_generation}
            $p \gets P(r)$ \tcp{Get respective prompt for relation $r$}
            $M_r \gets \text{LLM}(Q, B, p, N)$ \tcp{Generate $N$ mutations for relation $r$}
            $M \gets M \cup M_r$ \tcp{Add mutations to set} \label{line:end_mt_generation}

            \tcp{------------------ Mutation Verification ------------------}
            \ForEach{$m \in M_r$} {
                $f_m \gets \text{VerifyFactByLLM}(m, r)$ \tcp{Returns one of [$\Yes$, $\No$, $\NotSure$]} \label{line:mt_verification}

                \tcp{------------------ Hallucination Evaluation/Scoring ------------------}
                $S_m \gets S(r, f_m)$ \tcp{Compute hallucination score for mutation $m$} \label{line:start_mt_score}
                $S_{\text{total}} \gets S_{\text{total}} + S_m$ \tcp{Update total hallucination score}
            }
        }
        $S_{QB} \gets \frac{S_{total}}{|M|}$ \tcp{Scale between $[0, 1]$} \label{line:end_mt_score}
        
        \KwRet{$S_{QB}, M$} \tcp{Return total score and set of mutations}
    }
\end{algorithm}

%% file: main.bbl

\begin{thebibliography}{54}


\ifx \showCODEN    \undefined \def \showCODEN     #1{\unskip}     \fi
\ifx \showISBNx    \undefined \def \showISBNx     #1{\unskip}     \fi
\ifx \showISBNxiii \undefined \def \showISBNxiii  #1{\unskip}     \fi
\ifx \showISSN     \undefined \def \showISSN      #1{\unskip}     \fi
\ifx \showLCCN     \undefined \def \showLCCN      #1{\unskip}     \fi
\ifx \shownote     \undefined \def \shownote      #1{#1}          \fi
\ifx \showarticletitle \undefined \def \showarticletitle #1{#1}   \fi
\ifx \showURL      \undefined \def \showURL       {\relax}        \fi
\providecommand\bibfield[2]{#2}
\providecommand\bibinfo[2]{#2}
\providecommand\natexlab[1]{#1}
\providecommand\showeprint[2][]{arXiv:#2}

\bibitem[cha(2022)]%
        {chatgptintro}
 \bibinfo{year}{2022}\natexlab{}.
\newblock \bibinfo{booktitle}{\emph{OpenAI ChatGPT}}.
\newblock
\urldef\tempurl%
\url{https://openai.com/index/chatgpt/}
\showURL{%
\tempurl}


\bibitem[Abboud et~al\mbox{.}(2020)]%
        {abboud2020boxe}
\bibfield{author}{\bibinfo{person}{Ralph Abboud}, \bibinfo{person}{Ismail Ceylan}, \bibinfo{person}{Thomas Lukasiewicz}, {and} \bibinfo{person}{Tommaso Salvatori}.} \bibinfo{year}{2020}\natexlab{}.
\newblock \showarticletitle{Boxe: A box embedding model for knowledge base completion}.
\newblock \bibinfo{journal}{\emph{Advances in Neural Information Processing Systems}}  \bibinfo{volume}{33} (\bibinfo{year}{2020}), \bibinfo{pages}{9649--9661}.
\newblock


\bibitem[Agrawal et~al\mbox{.}(2023)]%
        {agrawal2023language}
\bibfield{author}{\bibinfo{person}{Ayush Agrawal}, \bibinfo{person}{Mirac Suzgun}, \bibinfo{person}{Lester Mackey}, {and} \bibinfo{person}{Adam~Tauman Kalai}.} \bibinfo{year}{2023}\natexlab{}.
\newblock \showarticletitle{Do Language Models Know When They're Hallucinating References?}
\newblock \bibinfo{journal}{\emph{arXiv preprint arXiv:2305.18248}} (\bibinfo{year}{2023}).
\newblock
\href{https://doi.org/10.48550/arXiv.2305.18248}{doi:\nolinkurl{10.48550/arXiv.2305.18248}}


\bibitem[Asher and Bhar(2024)]%
        {asher2024strong}
\bibfield{author}{\bibinfo{person}{Nicholas Asher} {and} \bibinfo{person}{Swarnadeep Bhar}.} \bibinfo{year}{2024}\natexlab{}.
\newblock \showarticletitle{Strong hallucinations from negation and how to fix them}.
\newblock \bibinfo{journal}{\emph{arXiv preprint arXiv:2402.10543}} (\bibinfo{year}{2024}).
\newblock
\href{https://doi.org/10.48550/arXiv.2402.10543}{doi:\nolinkurl{10.48550/arXiv.2402.10543}}


\bibitem[Atanasova et~al\mbox{.}(2020)]%
        {Atanasova2020GeneratingFC}
\bibfield{author}{\bibinfo{person}{Pepa Atanasova}, \bibinfo{person}{Jakob~Grue Simonsen}, \bibinfo{person}{Christina Lioma}, {and} \bibinfo{person}{Isabelle Augenstein}.} \bibinfo{year}{2020}\natexlab{}.
\newblock \showarticletitle{Generating Fact Checking Explanations}. In \bibinfo{booktitle}{\emph{Annual Meeting of the Association for Computational Linguistics}}.
\newblock
\urldef\tempurl%
\url{https://api.semanticscholar.org/CorpusID:215744944}
\showURL{%
\tempurl}


\bibitem[Augenstein et~al\mbox{.}(2019)]%
        {augenstein2019multifc}
\bibfield{author}{\bibinfo{person}{Isabelle Augenstein}, \bibinfo{person}{Christina Lioma}, \bibinfo{person}{Dongsheng Wang}, \bibinfo{person}{Lucas~Chaves Lima}, \bibinfo{person}{Casper Hansen}, \bibinfo{person}{Christian Hansen}, {and} \bibinfo{person}{Jakob~Grue Simonsen}.} \bibinfo{year}{2019}\natexlab{}.
\newblock \showarticletitle{MultiFC: A real-world multi-domain dataset for evidence-based fact checking of claims}.
\newblock \bibinfo{journal}{\emph{arXiv preprint arXiv:1909.03242}} (\bibinfo{year}{2019}).
\newblock
\href{https://doi.org/10.48550/arXiv.1909.03242}{doi:\nolinkurl{10.48550/arXiv.1909.03242}}


\bibitem[Chen et~al\mbox{.}(2023)]%
        {chen2023complex}
\bibfield{author}{\bibinfo{person}{Jifan Chen}, \bibinfo{person}{Grace Kim}, \bibinfo{person}{Aniruddh Sriram}, \bibinfo{person}{Greg Durrett}, {and} \bibinfo{person}{Eunsol Choi}.} \bibinfo{year}{2023}\natexlab{}.
\newblock \showarticletitle{Complex claim verification with evidence retrieved in the wild}.
\newblock \bibinfo{journal}{\emph{arXiv preprint arXiv:2305.11859}} (\bibinfo{year}{2023}).
\newblock
\href{https://doi.org/10.48550/arXiv.2305.11859}{doi:\nolinkurl{10.48550/arXiv.2305.11859}}


\bibitem[Chen et~al\mbox{.}(2018)]%
        {chen2018metamorphic}
\bibfield{author}{\bibinfo{person}{Tsong~Yueh Chen}, \bibinfo{person}{Fei-Ching Kuo}, \bibinfo{person}{Huai Liu}, \bibinfo{person}{Pak-Lok Poon}, \bibinfo{person}{Dave Towey}, \bibinfo{person}{TH Tse}, {and} \bibinfo{person}{Zhi~Quan Zhou}.} \bibinfo{year}{2018}\natexlab{}.
\newblock \showarticletitle{Metamorphic testing: A review of challenges and opportunities}.
\newblock \bibinfo{journal}{\emph{ACM Computing Surveys (CSUR)}} \bibinfo{volume}{51}, \bibinfo{number}{1} (\bibinfo{year}{2018}), \bibinfo{pages}{1--27}.
\newblock


\bibitem[Chern et~al\mbox{.}(2023)]%
        {chern2023factool}
\bibfield{author}{\bibinfo{person}{I Chern}, \bibinfo{person}{Steffi Chern}, \bibinfo{person}{Shiqi Chen}, \bibinfo{person}{Weizhe Yuan}, \bibinfo{person}{Kehua Feng}, \bibinfo{person}{Chunting Zhou}, \bibinfo{person}{Junxian He}, \bibinfo{person}{Graham Neubig}, \bibinfo{person}{Pengfei Liu}, {et~al\mbox{.}}} \bibinfo{year}{2023}\natexlab{}.
\newblock \showarticletitle{FacTool: Factuality Detection in Generative AI--A Tool Augmented Framework for Multi-Task and Multi-Domain Scenarios}.
\newblock \bibinfo{journal}{\emph{arXiv preprint arXiv:2307.13528}} (\bibinfo{year}{2023}).
\newblock
\href{https://doi.org/10.48550/arXiv.2307.13528}{doi:\nolinkurl{10.48550/arXiv.2307.13528}}


\bibitem[Cohen et~al\mbox{.}(2023)]%
        {cohen2023lm}
\bibfield{author}{\bibinfo{person}{Roi Cohen}, \bibinfo{person}{May Hamri}, \bibinfo{person}{Mor Geva}, {and} \bibinfo{person}{Amir Globerson}.} \bibinfo{year}{2023}\natexlab{}.
\newblock \showarticletitle{Lm vs lm: Detecting factual errors via cross examination}.
\newblock \bibinfo{journal}{\emph{arXiv preprint arXiv:2305.13281}} (\bibinfo{year}{2023}).
\newblock
\href{https://doi.org/10.48550/arXiv.2305.13281}{doi:\nolinkurl{10.48550/arXiv.2305.13281}}


\bibitem[Dhuliawala et~al\mbox{.}(2024)]%
        {dhuliawala2023chain}
\bibfield{author}{\bibinfo{person}{Shehzaad Dhuliawala}, \bibinfo{person}{Mojtaba Komeili}, \bibinfo{person}{Jing Xu}, \bibinfo{person}{Roberta Raileanu}, \bibinfo{person}{Xian Li}, \bibinfo{person}{Asli Celikyilmaz}, {and} \bibinfo{person}{Jason Weston}.} \bibinfo{year}{2024}\natexlab{}.
\newblock \showarticletitle{Chain-of-Verification Reduces Hallucination in Large Language Models}. In \bibinfo{booktitle}{\emph{Findings of the Association for Computational Linguistics ACL 2024}}, \bibfield{editor}{\bibinfo{person}{Lun-Wei Ku}, \bibinfo{person}{Andre Martins}, {and} \bibinfo{person}{Vivek Srikumar}} (Eds.). \bibinfo{publisher}{Association for Computational Linguistics}, \bibinfo{address}{Bangkok, Thailand and virtual meeting}, \bibinfo{pages}{3563--3578}.
\newblock
\urldef\tempurl%
\url{https://aclanthology.org/2024.findings-acl.212}
\showURL{%
\tempurl}


\bibitem[Foster et~al\mbox{.}(2025)]%
        {foster2025mutation}
\bibfield{author}{\bibinfo{person}{Christopher Foster}, \bibinfo{person}{Abhishek Gulati}, \bibinfo{person}{Mark Harman}, \bibinfo{person}{Inna Harper}, \bibinfo{person}{Ke Mao}, \bibinfo{person}{Jillian Ritchey}, \bibinfo{person}{Herv{\'e} Robert}, {and} \bibinfo{person}{Shubho Sengupta}.} \bibinfo{year}{2025}\natexlab{}.
\newblock \showarticletitle{Mutation-Guided LLM-based Test Generation at Meta}.
\newblock \bibinfo{journal}{\emph{arXiv preprint arXiv:2501.12862}} (\bibinfo{year}{2025}).
\newblock
\href{https://doi.org/10.48550/arXiv.2501.12862}{doi:\nolinkurl{10.48550/arXiv.2501.12862}}


\bibitem[Galitsky(2023)]%
        {202307.1723}
\bibfield{author}{\bibinfo{person}{Boris~A. Galitsky}.} \bibinfo{year}{2023}\natexlab{}.
\newblock \showarticletitle{Truth-O-Meter: Collaborating with LLM in Fighting its Hallucinations}.
\newblock \bibinfo{journal}{\emph{Preprints}} (\bibinfo{date}{July} \bibinfo{year}{2023}).
\newblock
\href{https://doi.org/10.20944/preprints202307.1723.v1}{doi:\nolinkurl{10.20944/preprints202307.1723.v1}}


\bibitem[Guo et~al\mbox{.}(2022)]%
        {guo2022survey}
\bibfield{author}{\bibinfo{person}{Zhijiang Guo}, \bibinfo{person}{Michael Schlichtkrull}, {and} \bibinfo{person}{Andreas Vlachos}.} \bibinfo{year}{2022}\natexlab{}.
\newblock \showarticletitle{A survey on automated fact-checking}.
\newblock \bibinfo{journal}{\emph{Transactions of the Association for Computational Linguistics}}  \bibinfo{volume}{10} (\bibinfo{year}{2022}), \bibinfo{pages}{178--206}.
\newblock


\bibitem[Hanselowski et~al\mbox{.}(2019)]%
        {Hanselowski2019ARA}
\bibfield{author}{\bibinfo{person}{Andreas Hanselowski}, \bibinfo{person}{Christian Stab}, \bibinfo{person}{Claudia Schulz}, \bibinfo{person}{Zile Li}, {and} \bibinfo{person}{Iryna Gurevych}.} \bibinfo{year}{2019}\natexlab{}.
\newblock \showarticletitle{A Richly Annotated Corpus for Different Tasks in Automated Fact-Checking}.
\newblock \bibinfo{journal}{\emph{ArXiv}}  \bibinfo{volume}{abs/1911.01214} (\bibinfo{year}{2019}).
\newblock
\urldef\tempurl%
\url{https://api.semanticscholar.org/CorpusID:207779874}
\showURL{%
\tempurl}


\bibitem[Huang et~al\mbox{.}(2023)]%
        {huang2023survey}
\bibfield{author}{\bibinfo{person}{Lei Huang}, \bibinfo{person}{Weijiang Yu}, \bibinfo{person}{Weitao Ma}, \bibinfo{person}{Weihong Zhong}, \bibinfo{person}{Zhangyin Feng}, \bibinfo{person}{Haotian Wang}, \bibinfo{person}{Qianglong Chen}, \bibinfo{person}{Weihua Peng}, \bibinfo{person}{Xiaocheng Feng}, \bibinfo{person}{Bing Qin}, {et~al\mbox{.}}} \bibinfo{year}{2023}\natexlab{}.
\newblock \showarticletitle{A survey on hallucination in large language models: Principles, taxonomy, challenges, and open questions}.
\newblock \bibinfo{journal}{\emph{arXiv preprint arXiv:2311.05232}} (\bibinfo{year}{2023}).
\newblock
\href{https://doi.org/10.1145/3703155}{doi:\nolinkurl{10.1145/3703155}}


\bibitem[Huo et~al\mbox{.}(2023)]%
        {huo2023retrieving}
\bibfield{author}{\bibinfo{person}{Siqing Huo}, \bibinfo{person}{Negar Arabzadeh}, {and} \bibinfo{person}{Charles~LA Clarke}.} \bibinfo{year}{2023}\natexlab{}.
\newblock \showarticletitle{Retrieving supporting evidence for llms generated answers}.
\newblock \bibinfo{journal}{\emph{arXiv preprint arXiv:2306.13781}} (\bibinfo{year}{2023}).
\newblock
\href{https://doi.org/10.48550/arXiv.2306.13781}{doi:\nolinkurl{10.48550/arXiv.2306.13781}}


\bibitem[Hyun et~al\mbox{.}(2024)]%
        {hyun2024metal}
\bibfield{author}{\bibinfo{person}{Sangwon Hyun}, \bibinfo{person}{Mingyu Guo}, {and} \bibinfo{person}{M.~Ali Babar}.} \bibinfo{year}{2024}\natexlab{}.
\newblock \showarticletitle{METAL: Metamorphic Testing Framework for Analyzing Large-Language Model Qualities}. In \bibinfo{booktitle}{\emph{2024 IEEE Conference on Software Testing, Verification and Validation (ICST)}}. \bibinfo{pages}{117--128}.
\newblock
\href{https://doi.org/10.1109/ICST60714.2024.00019}{doi:\nolinkurl{10.1109/ICST60714.2024.00019}}


\bibitem[Ji et~al\mbox{.}(2023)]%
        {ji2023survey}
\bibfield{author}{\bibinfo{person}{Ziwei Ji}, \bibinfo{person}{Nayeon Lee}, \bibinfo{person}{Rita Frieske}, \bibinfo{person}{Tiezheng Yu}, \bibinfo{person}{Dan Su}, \bibinfo{person}{Yan Xu}, \bibinfo{person}{Etsuko Ishii}, \bibinfo{person}{Ye~Jin Bang}, \bibinfo{person}{Andrea Madotto}, {and} \bibinfo{person}{Pascale Fung}.} \bibinfo{year}{2023}\natexlab{}.
\newblock \showarticletitle{Survey of hallucination in natural language generation}.
\newblock \bibinfo{journal}{\emph{Comput. Surveys}} \bibinfo{volume}{55}, \bibinfo{number}{12} (\bibinfo{year}{2023}), \bibinfo{pages}{1--38}.
\newblock


\bibitem[Kadavath et~al\mbox{.}(2022)]%
        {kadavath2022language}
\bibfield{author}{\bibinfo{person}{Saurav Kadavath}, \bibinfo{person}{Tom Conerly}, \bibinfo{person}{Amanda Askell}, \bibinfo{person}{Tom Henighan}, \bibinfo{person}{Dawn Drain}, \bibinfo{person}{Ethan Perez}, \bibinfo{person}{Nicholas Schiefer}, \bibinfo{person}{Zac Hatfield-Dodds}, \bibinfo{person}{Nova DasSarma}, \bibinfo{person}{Eli Tran-Johnson}, {et~al\mbox{.}}} \bibinfo{year}{2022}\natexlab{}.
\newblock \showarticletitle{Language models (mostly) know what they know}.
\newblock \bibinfo{journal}{\emph{arXiv preprint arXiv:2207.05221}} (\bibinfo{year}{2022}).
\newblock
\href{https://doi.org/10.48550/arXiv.2207.05221}{doi:\nolinkurl{10.48550/arXiv.2207.05221}}


\bibitem[Kasai et~al\mbox{.}(2024)]%
        {kasai2024realtime}
\bibfield{author}{\bibinfo{person}{Jungo Kasai}, \bibinfo{person}{Keisuke Sakaguchi}, \bibinfo{person}{Ronan Le~Bras}, \bibinfo{person}{Akari Asai}, \bibinfo{person}{Xinyan Yu}, \bibinfo{person}{Dragomir Radev}, \bibinfo{person}{Noah~A Smith}, \bibinfo{person}{Yejin Choi}, \bibinfo{person}{Kentaro Inui}, {et~al\mbox{.}}} \bibinfo{year}{2024}\natexlab{}.
\newblock \showarticletitle{REALTIME QA: what's the answer right now?}
\newblock \bibinfo{journal}{\emph{Advances in Neural Information Processing Systems}}  \bibinfo{volume}{36} (\bibinfo{year}{2024}).
\newblock


\bibitem[Kry{\'s}ci{\'n}ski et~al\mbox{.}(2019)]%
        {kryscinski2019evaluating}
\bibfield{author}{\bibinfo{person}{Wojciech Kry{\'s}ci{\'n}ski}, \bibinfo{person}{Bryan McCann}, \bibinfo{person}{Caiming Xiong}, {and} \bibinfo{person}{Richard Socher}.} \bibinfo{year}{2019}\natexlab{}.
\newblock \showarticletitle{Evaluating the factual consistency of abstractive text summarization}.
\newblock \bibinfo{journal}{\emph{arXiv preprint arXiv:1910.12840}} (\bibinfo{year}{2019}).
\newblock
\href{https://doi.org/10.48550/arXiv.1910.12840}{doi:\nolinkurl{10.48550/arXiv.1910.12840}}


\bibitem[Lai et~al\mbox{.}(2022)]%
        {lai2022exploration}
\bibfield{author}{\bibinfo{person}{Vivian Lai}, \bibinfo{person}{Alison Smith-Renner}, \bibinfo{person}{Ke Zhang}, \bibinfo{person}{Ruijia Cheng}, \bibinfo{person}{Wenjuan Zhang}, \bibinfo{person}{Joel Tetreault}, {and} \bibinfo{person}{Alejandro Jaimes}.} \bibinfo{year}{2022}\natexlab{}.
\newblock \showarticletitle{An exploration of post-editing effectiveness in text summarization}.
\newblock \bibinfo{journal}{\emph{arXiv preprint arXiv:2206.06383}} (\bibinfo{year}{2022}).
\newblock
\href{https://doi.org/10.48550/arXiv.2206.06383}{doi:\nolinkurl{10.48550/arXiv.2206.06383}}


\bibitem[Li et~al\mbox{.}(2023)]%
        {li2023halueval}
\bibfield{author}{\bibinfo{person}{Junyi Li}, \bibinfo{person}{Xiaoxue Cheng}, \bibinfo{person}{Wayne~Xin Zhao}, \bibinfo{person}{Jian-Yun Nie}, {and} \bibinfo{person}{Ji-Rong Wen}.} \bibinfo{year}{2023}\natexlab{}.
\newblock \showarticletitle{Halueval: A large-scale hallucination evaluation benchmark for large language models}.
\newblock \bibinfo{journal}{\emph{arXiv preprint arXiv:2305.11747}} (\bibinfo{year}{2023}).
\newblock
\href{https://doi.org/10.48550/arXiv.2305.11747}{doi:\nolinkurl{10.48550/arXiv.2305.11747}}


\bibitem[Li et~al\mbox{.}(2024)]%
        {li2024halluvault}
\bibfield{author}{\bibinfo{person}{Ningke Li}, \bibinfo{person}{Yuekang Li}, \bibinfo{person}{Yi Liu}, \bibinfo{person}{Ling Shi}, \bibinfo{person}{Kailong Wang}, {and} \bibinfo{person}{Haoyu Wang}.} \bibinfo{year}{2024}\natexlab{}.
\newblock \showarticletitle{HalluVault: A Novel Logic Programming-aided Metamorphic Testing Framework for Detecting Fact-Conflicting Hallucinations in Large Language Models}.
\newblock \bibinfo{journal}{\emph{arXiv preprint arXiv:2405.00648}} (\bibinfo{year}{2024}).
\newblock
\href{https://doi.org/10.48550/arXiv.2405.00648}{doi:\nolinkurl{10.48550/arXiv.2405.00648}}


\bibitem[Li et~al\mbox{.}(2022)]%
        {li2022faithfulness}
\bibfield{author}{\bibinfo{person}{Wei Li}, \bibinfo{person}{Wenhao Wu}, \bibinfo{person}{Moye Chen}, \bibinfo{person}{Jiachen Liu}, \bibinfo{person}{Xinyan Xiao}, {and} \bibinfo{person}{Hua Wu}.} \bibinfo{year}{2022}\natexlab{}.
\newblock \showarticletitle{Faithfulness in natural language generation: A systematic survey of analysis, evaluation and optimization methods}.
\newblock \bibinfo{journal}{\emph{arXiv preprint arXiv:2203.05227}} (\bibinfo{year}{2022}).
\newblock
\href{https://doi.org/10.48550/arXiv.2203.05227}{doi:\nolinkurl{10.48550/arXiv.2203.05227}}


\bibitem[Liang et~al\mbox{.}(2024)]%
        {liang2024survey}
\bibfield{author}{\bibinfo{person}{Ke Liang}, \bibinfo{person}{Lingyuan Meng}, \bibinfo{person}{Meng Liu}, \bibinfo{person}{Yue Liu}, \bibinfo{person}{Wenxuan Tu}, \bibinfo{person}{Siwei Wang}, \bibinfo{person}{Sihang Zhou}, \bibinfo{person}{Xinwang Liu}, \bibinfo{person}{Fuchun Sun}, {and} \bibinfo{person}{Kunlun He}.} \bibinfo{year}{2024}\natexlab{}.
\newblock \showarticletitle{A survey of knowledge graph reasoning on graph types: Static, dynamic, and multi-modal}.
\newblock \bibinfo{journal}{\emph{IEEE Transactions on Pattern Analysis and Machine Intelligence}} (\bibinfo{year}{2024}).
\newblock


\bibitem[Lin et~al\mbox{.}(2021)]%
        {lin2021truthfulqa}
\bibfield{author}{\bibinfo{person}{Stephanie Lin}, \bibinfo{person}{Jacob Hilton}, {and} \bibinfo{person}{Owain Evans}.} \bibinfo{year}{2021}\natexlab{}.
\newblock \showarticletitle{Truthfulqa: Measuring how models mimic human falsehoods}.
\newblock \bibinfo{journal}{\emph{arXiv preprint arXiv:2109.07958}} (\bibinfo{year}{2021}).
\newblock
\href{https://doi.org/10.48550/arXiv.2109.07958}{doi:\nolinkurl{10.48550/arXiv.2109.07958}}


\bibitem[Luo et~al\mbox{.}(2023)]%
        {luo2023chatgpt}
\bibfield{author}{\bibinfo{person}{Zheheng Luo}, \bibinfo{person}{Qianqian Xie}, {and} \bibinfo{person}{Sophia Ananiadou}.} \bibinfo{year}{2023}\natexlab{}.
\newblock \showarticletitle{Chatgpt as a factual inconsistency evaluator for text summarization}.
\newblock \bibinfo{journal}{\emph{arXiv preprint arXiv:2303.15621}} (\bibinfo{year}{2023}).
\newblock
\href{https://doi.org/10.48550/arXiv.2303.15621}{doi:\nolinkurl{10.48550/arXiv.2303.15621}}


\bibitem[Manakul et~al\mbox{.}(2023)]%
        {manakul2023selfcheckgpt}
\bibfield{author}{\bibinfo{person}{Potsawee Manakul}, \bibinfo{person}{Adian Liusie}, {and} \bibinfo{person}{Mark~JF Gales}.} \bibinfo{year}{2023}\natexlab{}.
\newblock \showarticletitle{Selfcheckgpt: Zero-resource black-box hallucination detection for generative large language models}.
\newblock \bibinfo{journal}{\emph{arXiv preprint arXiv:2303.08896}} (\bibinfo{year}{2023}).
\newblock
\href{https://doi.org/10.48550/arXiv.2303.08896}{doi:\nolinkurl{10.48550/arXiv.2303.08896}}


\bibitem[Min et~al\mbox{.}(2023)]%
        {min2023factscore}
\bibfield{author}{\bibinfo{person}{Sewon Min}, \bibinfo{person}{Kalpesh Krishna}, \bibinfo{person}{Xinxi Lyu}, \bibinfo{person}{Mike Lewis}, \bibinfo{person}{Wen-tau Yih}, \bibinfo{person}{Pang~Wei Koh}, \bibinfo{person}{Mohit Iyyer}, \bibinfo{person}{Luke Zettlemoyer}, {and} \bibinfo{person}{Hannaneh Hajishirzi}.} \bibinfo{year}{2023}\natexlab{}.
\newblock \showarticletitle{Factscore: Fine-grained atomic evaluation of factual precision in long form text generation}.
\newblock \bibinfo{journal}{\emph{arXiv preprint arXiv:2305.14251}} (\bibinfo{year}{2023}).
\newblock
\href{https://doi.org/10.48550/arXiv.2305.14251}{doi:\nolinkurl{10.48550/arXiv.2305.14251}}


\bibitem[Moon et~al\mbox{.}(2019)]%
        {Moon2019OpenDialKGEC}
\bibfield{author}{\bibinfo{person}{Seungwhan Moon}, \bibinfo{person}{Pararth Shah}, \bibinfo{person}{Anuj Kumar}, {and} \bibinfo{person}{Rajen Subba}.} \bibinfo{year}{2019}\natexlab{}.
\newblock \showarticletitle{OpenDialKG: Explainable Conversational Reasoning with Attention-based Walks over Knowledge Graphs}. In \bibinfo{booktitle}{\emph{Annual Meeting of the Association for Computational Linguistics}}.
\newblock
\urldef\tempurl%
\url{https://api.semanticscholar.org/CorpusID:196176000}
\showURL{%
\tempurl}


\bibitem[Moramarco et~al\mbox{.}(2021)]%
        {moramarco2021preliminary}
\bibfield{author}{\bibinfo{person}{Francesco Moramarco}, \bibinfo{person}{Alex~Papadopoulos Korfiatis}, \bibinfo{person}{Aleksandar Savkov}, {and} \bibinfo{person}{Ehud Reiter}.} \bibinfo{year}{2021}\natexlab{}.
\newblock \showarticletitle{A preliminary study on evaluating consultation notes with post-editing}.
\newblock \bibinfo{journal}{\emph{arXiv preprint arXiv:2104.04402}} (\bibinfo{year}{2021}).
\newblock
\href{https://doi.org/10.48550/arXiv.2104.04402}{doi:\nolinkurl{10.48550/arXiv.2104.04402}}


\bibitem[OpenAI et~al\mbox{.}(2023)]%
        {openai2023gpt}
\bibfield{author}{\bibinfo{person}{R OpenAI} {et~al\mbox{.}}} \bibinfo{year}{2023}\natexlab{}.
\newblock \showarticletitle{GPT-4 technical report}.
\newblock \bibinfo{journal}{\emph{ArXiv}}  \bibinfo{volume}{2303} (\bibinfo{year}{2023}), \bibinfo{pages}{08774}.
\newblock
\href{https://doi.org/10.48550/arXiv.2303.08774}{doi:\nolinkurl{10.48550/arXiv.2303.08774}}


\bibitem[Ouyang et~al\mbox{.}(2025)]%
        {Ouyang2025}
\bibfield{author}{\bibinfo{person}{Shuyin Ouyang}, \bibinfo{person}{Jie~M. Zhang}, \bibinfo{person}{Mark Harman}, {and} \bibinfo{person}{Meng Wang}.} \bibinfo{year}{2025}\natexlab{}.
\newblock \showarticletitle{An Empirical Study of the Non-Determinism of ChatGPT in Code Generation}.
\newblock \bibinfo{journal}{\emph{ACM Trans. Softw. Eng. Methodol.}} \bibinfo{volume}{34}, \bibinfo{number}{2}, Article \bibinfo{articleno}{42} (\bibinfo{date}{Jan.} \bibinfo{year}{2025}), \bibinfo{numpages}{28}~pages.
\newblock
\showISSN{1049-331X}


\bibitem[Qin et~al\mbox{.}(2024)]%
        {qin2024large}
\bibfield{author}{\bibinfo{person}{Libo Qin}, \bibinfo{person}{Qiguang Chen}, \bibinfo{person}{Xiachong Feng}, \bibinfo{person}{Yang Wu}, \bibinfo{person}{Yongheng Zhang}, \bibinfo{person}{Yinghui Li}, \bibinfo{person}{Min Li}, \bibinfo{person}{Wanxiang Che}, {and} \bibinfo{person}{Philip~S Yu}.} \bibinfo{year}{2024}\natexlab{}.
\newblock \showarticletitle{Large language models meet nlp: A survey}.
\newblock \bibinfo{journal}{\emph{arXiv preprint arXiv:2405.12819}} (\bibinfo{year}{2024}).
\newblock
\href{https://doi.org/10.48550/arXiv.2405.12819}{doi:\nolinkurl{10.48550/arXiv.2405.12819}}


\bibitem[Ren and Leskovec(2020)]%
        {ren2020beta}
\bibfield{author}{\bibinfo{person}{Hongyu Ren} {and} \bibinfo{person}{Jure Leskovec}.} \bibinfo{year}{2020}\natexlab{}.
\newblock \showarticletitle{Beta embeddings for multi-hop logical reasoning in knowledge graphs}.
\newblock \bibinfo{journal}{\emph{Advances in Neural Information Processing Systems}}  \bibinfo{volume}{33} (\bibinfo{year}{2020}), \bibinfo{pages}{19716--19726}.
\newblock


\bibitem[See et~al\mbox{.}(2017)]%
        {see2017get}
\bibfield{author}{\bibinfo{person}{Abigail See}, \bibinfo{person}{Peter~J Liu}, {and} \bibinfo{person}{Christopher~D Manning}.} \bibinfo{year}{2017}\natexlab{}.
\newblock \showarticletitle{Get to the point: Summarization with pointer-generator networks}.
\newblock \bibinfo{journal}{\emph{arXiv preprint arXiv:1704.04368}} (\bibinfo{year}{2017}).
\newblock
\href{https://doi.org/10.48550/arXiv.1704.04368}{doi:\nolinkurl{10.48550/arXiv.1704.04368}}


\bibitem[Sun et~al\mbox{.}(2020)]%
        {sun2020automatic}
\bibfield{author}{\bibinfo{person}{Zeyu Sun}, \bibinfo{person}{Jie~M Zhang}, \bibinfo{person}{Mark Harman}, \bibinfo{person}{Mike Papadakis}, {and} \bibinfo{person}{Lu Zhang}.} \bibinfo{year}{2020}\natexlab{}.
\newblock \showarticletitle{Automatic testing and improvement of machine translation}. In \bibinfo{booktitle}{\emph{Proceedings of the ACM/IEEE 42nd international conference on software engineering}}. \bibinfo{pages}{974--985}.
\newblock


\bibitem[Tian et~al\mbox{.}(2022)]%
        {tian2022knowledge}
\bibfield{author}{\bibinfo{person}{Ling Tian}, \bibinfo{person}{Xue Zhou}, \bibinfo{person}{Yan-Ping Wu}, \bibinfo{person}{Wang-Tao Zhou}, \bibinfo{person}{Jin-Hao Zhang}, {and} \bibinfo{person}{Tian-Shu Zhang}.} \bibinfo{year}{2022}\natexlab{}.
\newblock \showarticletitle{Knowledge graph and knowledge reasoning: A systematic review}.
\newblock \bibinfo{journal}{\emph{Journal of Electronic Science and Technology}} \bibinfo{volume}{20}, \bibinfo{number}{2} (\bibinfo{year}{2022}), \bibinfo{pages}{100159}.
\newblock


\bibitem[Varshney et~al\mbox{.}(2024)]%
        {varshney2024investigating}
\bibfield{author}{\bibinfo{person}{Neeraj Varshney}, \bibinfo{person}{Satyam Raj}, \bibinfo{person}{Venkatesh Mishra}, \bibinfo{person}{Agneet Chatterjee}, \bibinfo{person}{Ritika Sarkar}, \bibinfo{person}{Amir Saeidi}, {and} \bibinfo{person}{Chitta Baral}.} \bibinfo{year}{2024}\natexlab{}.
\newblock \showarticletitle{Investigating and Addressing Hallucinations of LLMs in Tasks Involving Negation}.
\newblock \bibinfo{journal}{\emph{arXiv preprint arXiv:2406.05494}} (\bibinfo{year}{2024}).
\newblock
\href{https://doi.org/10.48550/arXiv.2406.05494}{doi:\nolinkurl{10.48550/arXiv.2406.05494}}


\bibitem[Varshney et~al\mbox{.}(2023)]%
        {varshney2023stitch}
\bibfield{author}{\bibinfo{person}{Neeraj Varshney}, \bibinfo{person}{Wenlin Yao}, \bibinfo{person}{Hongming Zhang}, \bibinfo{person}{Jianshu Chen}, {and} \bibinfo{person}{Dong Yu}.} \bibinfo{year}{2023}\natexlab{}.
\newblock \showarticletitle{A stitch in time saves nine: Detecting and mitigating hallucinations of llms by validating low-confidence generation}.
\newblock \bibinfo{journal}{\emph{arXiv preprint arXiv:2307.03987}} (\bibinfo{year}{2023}).
\newblock
\href{https://doi.org/10.48550/arXiv.2307.03987}{doi:\nolinkurl{10.48550/arXiv.2307.03987}}


\bibitem[Vu et~al\mbox{.}(2023)]%
        {vu2023freshllms}
\bibfield{author}{\bibinfo{person}{Tu Vu}, \bibinfo{person}{Mohit Iyyer}, \bibinfo{person}{Xuezhi Wang}, \bibinfo{person}{Noah Constant}, \bibinfo{person}{Jerry Wei}, \bibinfo{person}{Jason Wei}, \bibinfo{person}{Chris Tar}, \bibinfo{person}{Yun-Hsuan Sung}, \bibinfo{person}{Denny Zhou}, \bibinfo{person}{Quoc Le}, {et~al\mbox{.}}} \bibinfo{year}{2023}\natexlab{}.
\newblock \showarticletitle{Freshllms: Refreshing large language models with search engine augmentation}.
\newblock \bibinfo{journal}{\emph{arXiv preprint arXiv:2310.03214}} (\bibinfo{year}{2023}).
\newblock
\href{https://doi.org/10.48550/arXiv.2310.03214}{doi:\nolinkurl{10.48550/arXiv.2310.03214}}


\bibitem[Xiong et~al\mbox{.}(2023)]%
        {xiong2023can}
\bibfield{author}{\bibinfo{person}{Miao Xiong}, \bibinfo{person}{Zhiyuan Hu}, \bibinfo{person}{Xinyang Lu}, \bibinfo{person}{Yifei Li}, \bibinfo{person}{Jie Fu}, \bibinfo{person}{Junxian He}, {and} \bibinfo{person}{Bryan Hooi}.} \bibinfo{year}{2023}\natexlab{}.
\newblock \showarticletitle{Can llms express their uncertainty? an empirical evaluation of confidence elicitation in llms}.
\newblock \bibinfo{journal}{\emph{arXiv preprint arXiv:2306.13063}} (\bibinfo{year}{2023}).
\newblock
\href{https://doi.org/10.48550/arXiv.2306.13063}{doi:\nolinkurl{10.48550/arXiv.2306.13063}}


\bibitem[Xu et~al\mbox{.}(2024)]%
        {xu2024hallucination}
\bibfield{author}{\bibinfo{person}{Ziwei Xu}, \bibinfo{person}{Sanjay Jain}, {and} \bibinfo{person}{Mohan Kankanhalli}.} \bibinfo{year}{2024}\natexlab{}.
\newblock \showarticletitle{Hallucination is inevitable: An innate limitation of large language models}.
\newblock \bibinfo{journal}{\emph{arXiv preprint arXiv:2401.11817}} (\bibinfo{year}{2024}).
\newblock
\href{https://doi.org/10.48550/arXiv.2401.11817}{doi:\nolinkurl{10.48550/arXiv.2401.11817}}


\bibitem[Yang et~al\mbox{.}(2018)]%
        {yang2018hotpotqa}
\bibfield{author}{\bibinfo{person}{Zhilin Yang}, \bibinfo{person}{Peng Qi}, \bibinfo{person}{Saizheng Zhang}, \bibinfo{person}{Yoshua Bengio}, \bibinfo{person}{William~W Cohen}, \bibinfo{person}{Ruslan Salakhutdinov}, {and} \bibinfo{person}{Christopher~D Manning}.} \bibinfo{year}{2018}\natexlab{}.
\newblock \showarticletitle{HotpotQA: A dataset for diverse, explainable multi-hop question answering}.
\newblock \bibinfo{journal}{\emph{arXiv preprint arXiv:1809.09600}} (\bibinfo{year}{2018}).
\newblock
\href{https://doi.org/10.48550/arXiv.1809.09600}{doi:\nolinkurl{10.48550/arXiv.1809.09600}}


\bibitem[Yao et~al\mbox{.}(2023)]%
        {yao2023llm}
\bibfield{author}{\bibinfo{person}{Jia-Yu Yao}, \bibinfo{person}{Kun-Peng Ning}, \bibinfo{person}{Zhen-Hui Liu}, \bibinfo{person}{Mu-Nan Ning}, {and} \bibinfo{person}{Li Yuan}.} \bibinfo{year}{2023}\natexlab{}.
\newblock \showarticletitle{Llm lies: Hallucinations are not bugs, but features as adversarial examples}.
\newblock \bibinfo{journal}{\emph{arXiv preprint arXiv:2310.01469}} (\bibinfo{year}{2023}).
\newblock
\href{https://doi.org/10.48550/arXiv.2310.01469}{doi:\nolinkurl{10.48550/arXiv.2310.01469}}


\bibitem[Yao et~al\mbox{.}(2024)]%
        {yao2024survey}
\bibfield{author}{\bibinfo{person}{Yifan Yao}, \bibinfo{person}{Jinhao Duan}, \bibinfo{person}{Kaidi Xu}, \bibinfo{person}{Yuanfang Cai}, \bibinfo{person}{Zhibo Sun}, {and} \bibinfo{person}{Yue Zhang}.} \bibinfo{year}{2024}\natexlab{}.
\newblock \showarticletitle{A survey on large language model (llm) security and privacy: The good, the bad, and the ugly}.
\newblock \bibinfo{journal}{\emph{High-Confidence Computing}} (\bibinfo{year}{2024}), \bibinfo{pages}{100211}.
\newblock


\bibitem[Yu et~al\mbox{.}({[n.\,d.]})]%
        {yucan}
\bibfield{author}{\bibinfo{person}{Chan~Xing Yu}, \bibinfo{person}{Chan Si~Yu James}, {and} \bibinfo{person}{Poh Hui-Li~Phyllis David}.} \bibinfo{year}{[n.\,d.]}\natexlab{}.
\newblock \showarticletitle{CAN LLMS HAVE A FEVER? INVESTIGATING THE EFFECTS OF TEMPERATURE ON LLM SECURITY}.
\newblock  (\bibinfo{year}{[n.\,d.]}).
\newblock


\bibitem[Zhang et~al\mbox{.}(2014)]%
        {zhang2014search}
\bibfield{author}{\bibinfo{person}{Jie Zhang}, \bibinfo{person}{Junjie Chen}, \bibinfo{person}{Dan Hao}, \bibinfo{person}{Yingfei Xiong}, \bibinfo{person}{Bing Xie}, \bibinfo{person}{Lu Zhang}, {and} \bibinfo{person}{Hong Mei}.} \bibinfo{year}{2014}\natexlab{}.
\newblock \showarticletitle{Search-based inference of polynomial metamorphic relations}. In \bibinfo{booktitle}{\emph{Proceedings of the 29th ACM/IEEE international conference on Automated software engineering}}. \bibinfo{pages}{701--712}.
\newblock


\bibitem[Zhang et~al\mbox{.}(2022)]%
        {9000651}
\bibfield{author}{\bibinfo{person}{Jie~M. Zhang}, \bibinfo{person}{Mark Harman}, \bibinfo{person}{Lei Ma}, {and} \bibinfo{person}{Yang Liu}.} \bibinfo{year}{2022}\natexlab{}.
\newblock \showarticletitle{Machine Learning Testing: Survey, Landscapes and Horizons}.
\newblock \bibinfo{journal}{\emph{IEEE Transactions on Software Engineering}} \bibinfo{volume}{48}, \bibinfo{number}{1} (\bibinfo{year}{2022}), \bibinfo{pages}{1--36}.
\newblock


\bibitem[Zhang et~al\mbox{.}(2023a)]%
        {zhang2023concepteva}
\bibfield{author}{\bibinfo{person}{Xiaoyu Zhang}, \bibinfo{person}{Jianping Li}, \bibinfo{person}{Po-Wei Chi}, \bibinfo{person}{Senthil Chandrasegaran}, {and} \bibinfo{person}{Kwan-Liu Ma}.} \bibinfo{year}{2023}\natexlab{a}.
\newblock \showarticletitle{ConceptEVA: concept-based interactive exploration and customization of document summaries}. In \bibinfo{booktitle}{\emph{Proceedings of the 2023 CHI Conference on Human Factors in Computing Systems}}. \bibinfo{pages}{1--16}.
\newblock


\bibitem[Zhang et~al\mbox{.}(2023b)]%
        {zhang2023siren}
\bibfield{author}{\bibinfo{person}{Yue Zhang}, \bibinfo{person}{Yafu Li}, \bibinfo{person}{Leyang Cui}, \bibinfo{person}{Deng Cai}, \bibinfo{person}{Lemao Liu}, \bibinfo{person}{Tingchen Fu}, \bibinfo{person}{Xinting Huang}, \bibinfo{person}{Enbo Zhao}, \bibinfo{person}{Yu Zhang}, \bibinfo{person}{Yulong Chen}, {et~al\mbox{.}}} \bibinfo{year}{2023}\natexlab{b}.
\newblock \showarticletitle{Siren's song in the AI ocean: a survey on hallucination in large language models}.
\newblock \bibinfo{journal}{\emph{arXiv preprint arXiv:2309.01219}} (\bibinfo{year}{2023}).
\newblock
\href{https://doi.org/10.48550/arXiv.2309.01219}{doi:\nolinkurl{10.48550/arXiv.2309.01219}}


\bibitem[Zhou et~al\mbox{.}(2019)]%
        {zhou2019completing}
\bibfield{author}{\bibinfo{person}{Zili Zhou}, \bibinfo{person}{Shaowu Liu}, \bibinfo{person}{Guandong Xu}, {and} \bibinfo{person}{Wu Zhang}.} \bibinfo{year}{2019}\natexlab{}.
\newblock \showarticletitle{On completing sparse knowledge base with transitive relation embedding}. In \bibinfo{booktitle}{\emph{Proceedings of the AAAI Conference on Artificial Intelligence}}, Vol.~\bibinfo{volume}{33}. \bibinfo{pages}{3125--3132}.
\newblock


\end{thebibliography}
